\documentclass{article}
\usepackage{arxiv}
\usepackage[utf8]{inputenc} 
\usepackage[T1]{fontenc}    
\usepackage{hyperref}       
\usepackage{url}            
\usepackage{booktabs}       
\usepackage{amsfonts}       
\usepackage{nicefrac}       
\usepackage{microtype}      
\usepackage{lipsum}
\usepackage{graphicx}
\usepackage{comment}
\usepackage{longtable} 
\usepackage{amsmath}

\usepackage{multirow}

\title{Does Medical Specialization of VLMs Enhance Discriminative Power?:\\
A Comprehensive Investigation through Feature Distribution Analysis}

\author{
 Keita Takeda \\
  Graduate School of Integrated Science and Technology\\
  Nagasaki University\\
  \texttt{ktakeda@nagasaki-u.ac.jp} \\
  \And
 Tomoya Sakai\\
  Graduate School of Integrated Science and Technology\\
  Nagasaki University\\
  \texttt{tsakai@nagasaki-u.ac.jp} \\
}

\begin{document}
\maketitle
\begin{abstract}
This study investigates the feature representations produced by publicly available open source medical vision-language models (VLMs).  
While medical VLMs are expected to capture diagnostically relevant features, their learned representations remain underexplored, and standard evaluations like classification accuracy do not fully reveal if they acquire truly discriminative, lesion-specific features. 
Understanding these representations is crucial for revealing medical image structures and improving downstream tasks in medical image analysis.
This study aims to investigate the feature distributions learned by medical VLMs and evaluate the impact of medical specialization. 
We analyze the feature distribution of multiple image modalities extracted by some representative medical VLMs across lesion classification datasets on multiple modalities. 
These distributions were compared them with non-medical VLMs to assess the domain-specific medical training. 
Our experiments showed that medical VLMs can extract discriminative features that are effective for medical classification tasks. Moreover, it was found that non-medical VLMs with recent improvement with contextual enrichment such as LLM2CLIP produce more refined feature representations. 
Our results imply that enhancing text encoder is more crucial than training intensively on medical images when developing medical VLMs. 
Notably, non-medical models are particularly vulnerable to biases introduced by overlaied text strings on images. 
These findings underscore the need for careful consideration on model selection according to downstream tasks besides potential risks in inference due to background biases such as textual information in images.
\end{abstract}


\begin{table}[tb]
\centering
\caption{Popular medical vision-language models (VLMs) and corresponding non-medical VLMs. Supported means the VLM is employed for experiment on this study. Backbone means the VLM employ the image encoder backbone. Target modality means the VLM is specialized the imaging modality. } \label{tab:famous_VLMs}
\hspace*{-0.7cm}
\begin{tabular}{cllll}
\toprule
 support & model name & backbone & target modality & dataset source \\ \midrule
&\multicolumn{4}{l}{Medical VLMs trained on contrastive learning manner} \\
\checkmark & BiomedCLIP~\cite{Zhang23} & ViT-B/16~\cite{dosovitskiy21} & Modality-agnostic & Paper figure \\
 & UniMedCLIP~\cite{Khattak24} & ViT-L/14~\cite{dosovitskiy21} & Modality-agnostic & Public datasets \\
\checkmark  & ConceptCLIP~\cite{nie2025explainable} & SigLIP~\cite{Zhai_2023_ICCV} & Modality-agnostic & Paper figure \\
\checkmark & MedSigLIP~\cite{sellergren2025medgemma} & SigLIP~\cite{Zhai_2023_ICCV} & Modality-agnostic & Paper figure and public datasets \\ 
 & CT-CLIP~\cite{Hamamci24} & CT-ViT~\cite{wang2022transformer} based & 3D-CT & Public datasets \\
 & Merlin~\cite{blankemeier2024merlin} & 3D ResNet152 & 3D-CT & Original \\ 
\checkmark & CXR-CLIP~\cite{you23} & SwinTransformer~\cite{Liu_2021_ICCV} & Chest X-ray & Public datasets \\
 & Mammo-CLIP~\cite{ghosh2024mammo} & EfficientNet~\cite{tan2019efficientnet} & Mammography & Public datasets \\
\checkmark & CONCH~\cite{Lu24} & CoCa~\cite{Yu22} based & Histopathology & Public datasets \\
 & ViLa-MIL~\cite{Shi_2024_CVPR} & CLIP~\cite{Radford21} & Histopathology & Public dataset \\
\checkmark & UNI~\cite{Khattak24} & DINOv2~\cite{Oquab23} (ViT-B/16~\cite{dosovitskiy21}) & Histopathology & Original \\
 & CHIEF~\cite{Wang24} & CTransPath~\cite{wang2022transformer} based & Histopathology & Public datasets \\
\checkmark & FLAIR~\cite{silva2025foundation} & ResNet50~\cite{He16} & Ophthalmology  & Public datasets \\
 & KeepFIT~\cite{wu2024mm} & ResNet50~\cite{He16} & Ophthalmology & Books and public datasets \\
 & VisionUite~\cite{Li24} & EVA02~\cite{Fang24} and CLIP~\cite{Radford21} based & Ophthalmology & Public datasets \\
 & CLIP-DR~\cite{yu2024clip} & ResNet50~\cite{He16} & Ophthalmology & Public datasets \\
\checkmark  & MONET~\cite{kim2024transparent} & ViT-L/14~\cite{dosovitskiy21} & Dermatology & Paper figure and public datasets \\
 & MAKE~\cite{Yan25make} & CLIP~\cite{Radford21} & Dermatology & Paper figure and others \\
 & DermLIP~\cite{yan2025multimodal} & ViT-L~\cite{dosovitskiy21} & Dermatology & Paper figure and public datasets \\ \midrule
&\multicolumn{4}{l}{Medical VLMs trained on instruction tuning manner} \\ 
\checkmark  & LLaVA-Med~\cite{Li23} & ViT-L/14-336~\cite{dosovitskiy21} & Modality-agnostic & Paper figure \\
\checkmark  & LLaVA-Med++~\cite{Xie24} & ViT-L/14-336~\cite{dosovitskiy21} & Modality-agnostic & Paper figure and public datasets \\
 & MedGemma~\cite{sellergren2025medgemma} & MedSigLIP~\cite{sellergren2025medgemma} & Modality-agnostic & Public datasets \\ \midrule
&\multicolumn{4}{l}{Corresponding or popular non-medical VLMs} \\
\checkmark   & LLaVA~\cite{Liu23, Liu24} & ViT-L/14-336~\cite{dosovitskiy21} & Non-medical & Public datasets \\
\checkmark  & EVA02~\cite{Fang24} & ViT-L/14-336~\cite{dosovitskiy21} & Non-medical & Public datasets \\
\checkmark   & LLM2CLIP~\cite{Huang24} & ViT-L/14-336~\cite{dosovitskiy21} & Non-medical & Public datasets \\ \bottomrule
\end{tabular}
\end{table}

\begin{table*}[tb]
    \centering
    \caption{Specifications of datasets used for experiment}
    \label{tab:dataset_specifications}
    \begin{tabular}{lll}
        \toprule
        Dataset Name & Modality & task \\
        \midrule
        Brain tumor classification~\cite{Bhuvaji20}    & MRI      & Brain tumor detection and classification \\ (referred to as Brain MR) & & \\
        SARS-COV-2 CTscan~\cite{soares20}    & CT       & COVID-19 diagnosis \\
        PneumoniaMNIST~\cite{medmnistv1,medmnistv2,Kermany18} & X-ray & Diagnosis of pneumonia \\
        Breast Ultrasound Images Dataset~\cite{AlDhabyani20}    & Ultrasound      & Breast tumor classification \\
        (referred to as BreastUS) & & \\
        BreakHis~\cite{Fabio16}    & Histopathology & Breast tumor classification \\
        HiCervix~\cite{Cai24}    & Cytology & Cervical cancer detection \\
        DeepDRiD~\cite{Liu22}    & Ophthalmology & Grading retinal images for \\
        & & diagnosis of Diabetic retinopathy \\
        ISIC~2019~\cite{Tschandl18,Codella18,Hernandez24} & Dermatology & Skin cancer detection\\
        \bottomrule
    \end{tabular}
\end{table*}

\section{Introduction}

Vision-Language Models (VLMs) are deep learning models trained on large-scale datasets to embed images and texts into their respective universal feature spaces~\cite{Zhang24}. While general VLMs have shown remarkable capabilities~\cite{Zhang24}, adapting them to specialized domains is crucial. Medical image analysis, in particular, has emerged as a significant application area~\cite{Chen24survey}. 
This field faces unique challenges, such as difficulties in acquiring large datasets due to limited cases and the critical need to analyze disease-specific image features. Consequently, as illustrated in Table.~\ref{tab:famous_VLMs}, numerous medical VLMs have been proposed in recent years to address these demands.

Understanding the universal features produced by these medical VLMs is paramount to assume their appropriate application to downstream tasks, especially in medical image analysis where explainability and interpretability are vital. 
However, research on the universal feature spaces of VLMs has often been limited to indirect evaluations, even in the natural image domain~\cite{Thrush22, Tu23}. 
Existing survey papers on medical VLMs~\cite{Shrestha23,Zhang24survey,Chen24survey} also tend to focus on evaluation via performance metrics like classification accuracy rather than re-evaluating the learned universal feature spaces. 
Therefore, a deeper investigation into what features are internally computed by these models and how their parameters contribute to feature extraction is essential.

To address this gap, our study analyzes the distribution of image features extracted by the image encoders of medical VLMs. 
We utilize eight classification datasets covering different imaging modalities. 
The universal image feature distributions extracted by representative medical VLMs shown in Table.~\ref{tab:famous_VLMs} are visualized using dimensionality reduction. 
These are compared with feature distributions from their non-medical counterparts as well as LLM2CLIP~\cite{Huang24} (which features an enhanced text encoder for contextual enrichment), to assess the impact of medical domain-specific training. 
Furthermore, since low-dimensional visualization may not fully capture a model's ability to discern discriminative features, we construct linear discriminators to evaluate the performance of each model in its original high-dimensional feature space. 

\section{Motivation of developping Medical VLMs}

The development of medical VLMs is driven by two primary motivations. 
The first is to achieve SoTA model performance. 
VLMs are adept at learning high-quality feature representations compared to conventional models~\cite{zhang2022contrastive, joulin2016learning, Radford21}, making it easier to outperform models trained without language labels~\cite{Radford21}. 
Indeed, several medical VLMs have been evaluated to have superior performance on benchmark datasets compared to non language models such as ResNet~\cite{He16} or SimCLR~\cite{chen2020simple, chen2020big}. 
A key byproduct of this is the facilitation of transfer learning for rare diseases~\cite{lu2025integrating}. 
Since VLMs are assumed to have learned sophisticated feature representations well-suited for a wide range of downstream tasks~\cite{Zhang23}, they are expected to achieve high performance even on tasks where non VLMs struggled to overfitting~\cite{Rodriguez25, lu2025integrating}. 

The second motivation is to advance medical image understanding through explanations via language. 
The diagnoses of medical image are still relies heavily on empirical knowledge of clinicians. 
While many DNN approaches have aimed to quantify and improve the empirical modeling, the subjective nature of these assessments makes them difficult to annotate, posing a challenge for traditional supervised learning. 
However, these empirical findings are often recorded by text in clinical reports~\cite{zhang2022contrastive}. 
Training VLMs with emprical descriptions as a supervision enables to quantify image features that correspond to these clinical knowledge~\cite{zhang2022contrastive}. 
This linkage between visual features and language also enhances model explainability.

\section{Data sources for training medical VLMs}
Medical VLMs requires vast amounts of data for their training. Current approaches draw from two primary data sources: aggregation of public datasets, and extraction of medical images from medical journal papers.

First, a common strategy is the aggregation of public datasets. 
While individual medical datasets are often too small for training large VLMs, combining them can yield a substantial volume of several million images~\cite{li2024gmai, Khattak24}. 
Popular portal sites include the cancer imaging archive (TCIA)~\cite{clark2013cancer}, which hosts a comprehensive collection of cancer-related images (MRI, CT, pathology) along with clinical metadata and reports. 
Other valuable resources are Zenodo~\cite{Zenodo}, which archives data used on experiment of research papers, and Kaggle~\cite{kaggle}, which hosts curated datasets for competitions. 

While aggregating public dataset allows us to build a large dataset covered for wide-range modalities, some applications requires feature representations specialized for narrower, specific modalities. 
As will be discussed in Section~\ref{s:MajorVLM}, the development of modality-specific datasets is advancing to build such specialized VLMs. 
Constructing modality-specific datasets is typically achieved by aggregating numerous smaller datasets around a core of one or more relatively large-scale datasets, which usually contain tens of thousands of images or more. 

Chest X-ray and fundus imaging are the modalities for which the construction of modality-specific datasets is particularly advanced. 
For chest X-ray data, two dataset, CheXpert~\cite{irvin2019chexpert} and MIMIC-CXR~\cite{johnson2024mimic}, are highly influential. 
CheXpert is a multi-label dataset created to assimulate the findings of radiologists. 
While multi-label datasets require special handling for standard classifiers, they are valuable for training foundation models as they facilitate the learning of image features that correspond to a wide range of linguistic expressions.

In the fundus imaging modality, the two most prominent datasets are EYEPACK~\cite{Eyepack} and AIROGS~\cite{AIROGS}. 
These are categorical datasets for the classification of diabetic retinopathy and glaucoma screening, respectively, and are substantially larger than other available datasets. 
A typical example of dataset aggregation in this modality is the FLAIR~\cite{Rodriguez25}, which Rodriguez et al. constructed by integrating $38$ public datasets. 

For other modalities, the aggregation of public datasets is less common. 
A significant factor is the limited number of publicly available datasets. 
Particularly for MR imaging, while numerous segmentation datasets have been proposed, categorical datasets are scarce. 
This is due to the primary focus on detection tasks within the MR domain. 
Similarly, for other modalities where detection tasks are predominant, such as CT, there are fewer proposals for VLMs. 
Likewise, for modality such as ultrasound, where datasets are inherently scarce, the lack of data for aggregation has also limited the proposal of VLMs.

While these public datasets offer high-quality clinical images and corresponding image findings, their total volume often falls short of the tens of millions of samples used to train general-purpose VLMs. 
Consequently, they are typically used in conjunction with other data sources. 

Another major data source is the vast repository of figures and captions within academic publications. 
Medical journals frequently feature clinical images, and by leveraging curated collections like PMC-OA~\cite{lin2023pmc} or employing large-scale web scraping~\cite{Zhang23, ruckert2024rocov2, subramanian2020medicat, baghbanzadeh2025advancing}, researchers have been able to construct datasets in the tens of millions~\cite{Zhang23, ruckert2024rocov2, baghbanzadeh2025advancing}. 
Medical papers often contains a wide range of typical and outlier cases with high-quality image captions. 
However, a significant drawback is the image quality itself. 
Publication formats are not optimized for medical imaging formats such as DICOM. 
High-resolution, high-bit-depth images are often compressed to meet file size constraints, which may result in the loss of important diagnostic features~\cite{liu2017current, koff2013evaluation, herath2025systematic}. 
Therefore, the development of medical VLMs often relies on a hybrid approach, balancing the high-quality of public datasets with the large-quantity of data extracted from literature.

\section{Popular Medical VLMs and remaining challenges}\label{s:MajorVLM}
Current research in medical VLMs is primarily advancing along two main directions: modality-agnostic and modality-specific approaches. 
The first, modality-agnostic medical VLMs, aim to capture image features common across general medical images, without specializing in a particular domain.
Table~\ref{tab:famous_VLMs} shows popular VLMs and corresponding non-medical VLMs.
For instance, Zhang \textit{et al.} developed BiomedCLIP~\cite{Zhang23} by training CLIP~\cite{Radford21} on PMC-15M, a dataset of $15$ million figure-caption pairs from PubMed articles. 
Li \textit{et al.} developped LLaVA-Med~\cite{Li23} by fine-tuning LLaVA~\cite{Liu23} on the dataset. 
The performance of instraction tuned models on tasks of visual question answering (VQA) has been further improved by leveraging even larger datasets, such as MedTrinity-25M~\cite{Xie24}. 
Khattak~\textit{et al.} also introduced UniMedCLIP, trained on UniMed, a $5.3$ million sample dataset aggregated from public medical sources~\cite{Khattak24}. 
Sellergren~\textit{et al.} developed MedGemma~\cite{sellergren2025medgemma} by fine-tuning Gemma on PMC-OA~\cite{lin2023pmc} and several public dataset.
These modality-agnostic medical VLMs are expected to have a wide range of applications. 

The other group, modality-specific medical VLMs, aims to learn feature extractors specialized for particular imaging modalities. 
By specializing in a specific imaging modality, these models enable end-to-end analysis for that modality, often with a single, tailored model. 
In histopathology, for example, foundation models such as CONCH~\cite{Lu24}, UNI~\cite{Chen24}, and CHIEF~\cite{Wang24} have been developed, typically utilizing large collections of pathology images from resources like The Cancer Genome Atlas (TCGA)~\cite{Weinstein13}. 
In ophthalmology, VisionUnite~\cite{Li24} has emerged as a notable model, employing a Large Language Model (LLM) as its text encoder to achieve context-aware image feature understanding based on clinical findings. 
Furthermore, to address the complexities of 3D medical data, Hamamci \textit{et al.} extended the 2D CLIP architecture to 3D for learning structural features from CT volumes~\cite{Hamamci24}. 

However, how to best leverage these foundation models remains a crucial problem. 
In practice, directly applying a foundation model to a wide range of tasks is uncommon. 
Instead, research and clinical utility are typically directed toward narrow downstream objectives, such as the fine-grained classification or localized detection of specific pathological lesions.
For these specialized applications, compact and highly discriminative feature representations are frequently the most effective.
By distilling necessary and sufficient feature representations from medical VLMs, which have been pre-trained on diverse datasets, researchers can develop task specific models that are both computationally efficient and robust to the domain shifts common in clinical environments.

Many existing approaches employ fine-tuning approach to build task-specific models by collecting relatively large datasets containing over a thousand images. 
However, it is often impossible to prepare a dataset large enough for effective fine-tuning in the medical image domain. 
To address these constraints, parameter-efficient fine-tuning (PEFT) and zero-shot classification have emerged as compelling alternatives. 
These methodologies practically assume that the VLM's latent space already encompasses a rich feature representations relevant to the downstream task. 
While recent benchmarks~\cite{Shrestha23,Zhang24survey,Chen24survey} have validated the discriminative capabilities of these approaches, often through linear probing, they primarily focus on improvements of performance metrics like accuracy or F1-score. 
Such evaluations may obscure whether the model is truly capturing lesion-specific semantics or merely relying on spurious correlations.
Consequently, a deeper investigation into the VLM's learned universal feature space is crucial for ensuring the reliability and explainability of their outputs in diverse medical image analysis applications.

\section{Discriminative capability analysis via visualization of image feature distributions}

This study aims to ascertain whether medical VLMs effectively acquire discriminative features from medical specialization. 
To test this hypothesis, we visualizes the feature distributions extracted by representative medical VLMs across different modalities as listed in Table~\ref{tab:famous_VLMs}.
For comparison, we included their corresponding non-medical counterparts and other popular non-medical VLMs to assess the effectiveness of domain-specific medical training.

We investigated the modality-specific behaviour of feature extraction by observing the feature distribution of the dataset across eight modalities. 
We collected public datasets for MRI , CT , X-ray , ultrasound , histopathology , cytology , ophthalmology , and dermatology.
These datasets are detailed in Table~\ref{tab:dataset_specifications}.
Input images were pre-processed using the official functions provided with each VLM's implementation.

We extended our analysis to evaluate the zero-shot transferability of modality-specific VLMs to domains outside their primary training distribution. 
Specifically, we applied modality specific VLMs to both closely related and functionally distinct modalities. 
This assessment is predicated on the hypothesis that VLMs trained on large-scale datasets of distinct modalities may develop superior feature representations that surpass those of models trained on smaller, domain-relevant datasets. 
By including seemingly unrelated modalities, we aim to verify the extent to which data scale and inherent feature complexity influence cross-modal representation learning.

To qualitatively assess the discriminative ability of the extracted image features, we reduced them to two dimensions using UMAP~\cite{Leland20} and visualized the resulting distributions. 
We employed the Python implementation provided by Sainburg \textit{et al.}~\cite{sainburg21}. 
As most VLMs are trained via contrastive learning to enhance cosine similarity, we selected the \texttt{cosine} metric for UMAP optimization. 
While we conducted an extensive grid search over the $n\_neighbors$ and $min\_dist$ hyperparameters and analyzed the resulting distributions across all configurations, we present only the most representative visualizations that effectively characterize the observed feature properties.

Furthermore, since visualization of reduced-dimensional features alone may not fully reflect the ability of VLMs to capture discriminative features, we constructed a linear discriminator to evaluate the performance in the original high-dimensional feature space. 
For training the image classifiers, $20\%$ of the data (from the official training set, or the combined training and validation set where applicable) was held out as the test set. 
The remaining data were used for training with five-fold cross-validation.

We employed a linear SVM from \texttt{scikit-learn}~\cite{Buitinck13} implementation as the discriminator. 
All feature vectors were standardized before training. 
The SVM hyperparameter $C$ was set as $C=t d_e/N$, where the optimal $t$ was determined via grid search. Here, $d_e$ denotes the number of dimension of the feature vector extracted by each model, and $N$ is the number of training samples. The parameter $t$ was selected from $\{1, 10, 100, 1000, 10000\}$ to maximize the F1-score on the validation data. These parameters were sufficient to saturate the F1-score on the training and validation data; fundamentally, the classification performance on the validation data remained roughly equivalent regardless of which of these hyperparameters was utilized.

Finally, to qualitatively assess which image regions contributed to the classification, a qualitative evaluation of model attention was conducted. 
After training the linear classifier, the attention maps were visualized using RISE~\cite{Petsiuk2018rise} for the complete classification pipeline (i.e., the VLM encoder followed by the trained linear SVM). 
The number of random mask generations, $N$, was set to $4096$.

\subsection{The VLMs can be expected to classify tumors}
\begin{figure}[tb]
    \begin{center}
    \includegraphics[width=\linewidth]{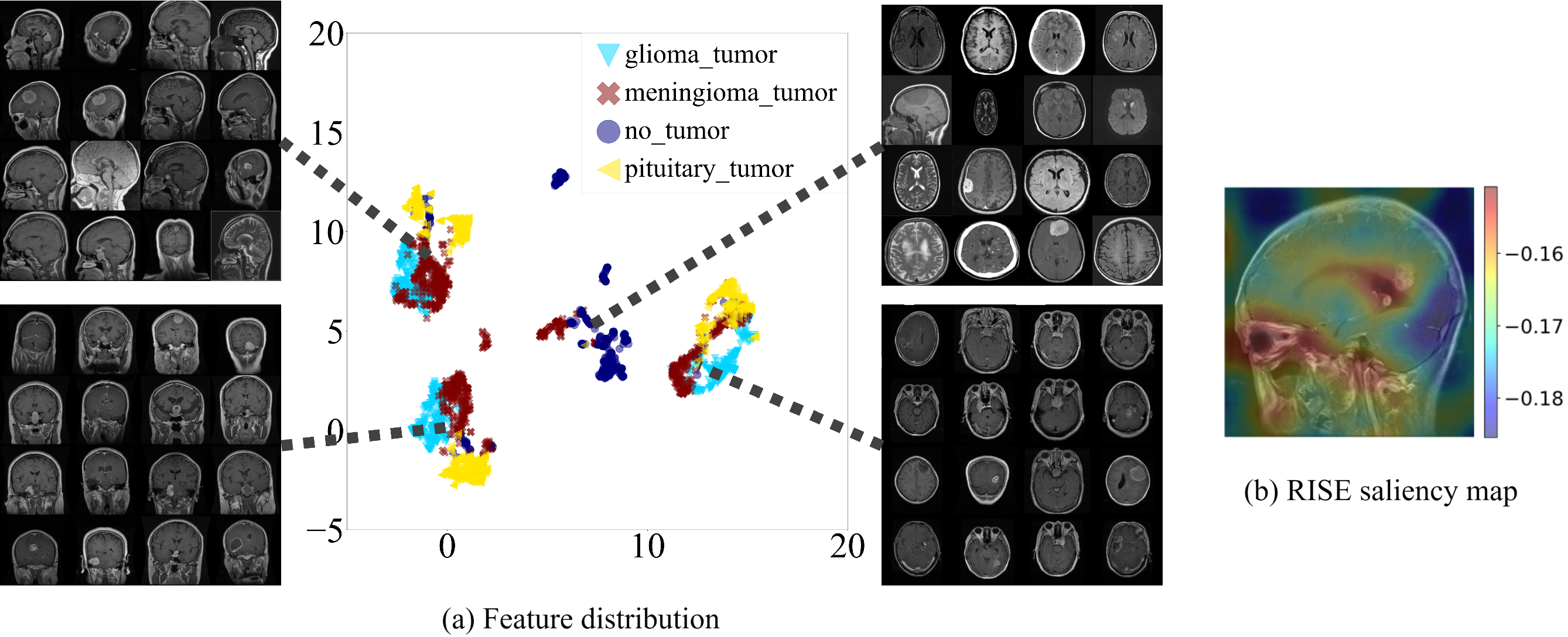}
    \caption{A exsample of feature distribution on Brain MR~\cite{Bhuvaji20} dataset and corresponding RISE saliency maps. 
    (a) Feature distribution and example images from the Brain MR~\cite{Bhuvaji20}, generated via LLM2CLIP~\cite{Huang24}. Three clusters are formed according to the anatomical plane, with the central cluster primarily composed of images without tumors. (b) A class activation map from the LLM2CLIP and linear SVM classifiers. The model focus on the lesion area and the orbital region.}
    \label{fig:FD_BrainMR}
    \end{center}
\end{figure}

The VLMs demonstrate the ability to acquire discriminative features related to the presence and malignancy of lesions. 
Figure~\ref{fig:FD_BrainMR}(a) shows the feature distribution for the brain MR tumor classification dataset via LLM2CLIP~\cite{Huang24}. 
Four distinct clusters are formed: three corresponding to the different anatomical planes, and a fourth primarily composed of images without tumors. 
Within these clusters, data points are discriminatively distributed regarding tumor type.

Furthermore, we constructed a linear discriminator and visualized its attention using RISE~\cite{Petsiuk2018rise}. 
The resulting attention maps, shown in Figure~\ref{fig:FD_BrainMR}(b).
A discriminator pay attention on the tumor regions for classification. 
This indicates that the VLMs successfully learned discriminative features for both the presence and malignancy of tumors.

These findings were consistent across nearly all datasets and VLMs evaluated, including the non-medical VLMs. 
Consequently, for the datasets examined in this study, we did not observe a significant advantage from specializing the models on medical images.

\subsection{Medical VLMs are affected by background biases}
\begin{figure}[tb]
    \begin{center}
    \includegraphics[width=\linewidth]{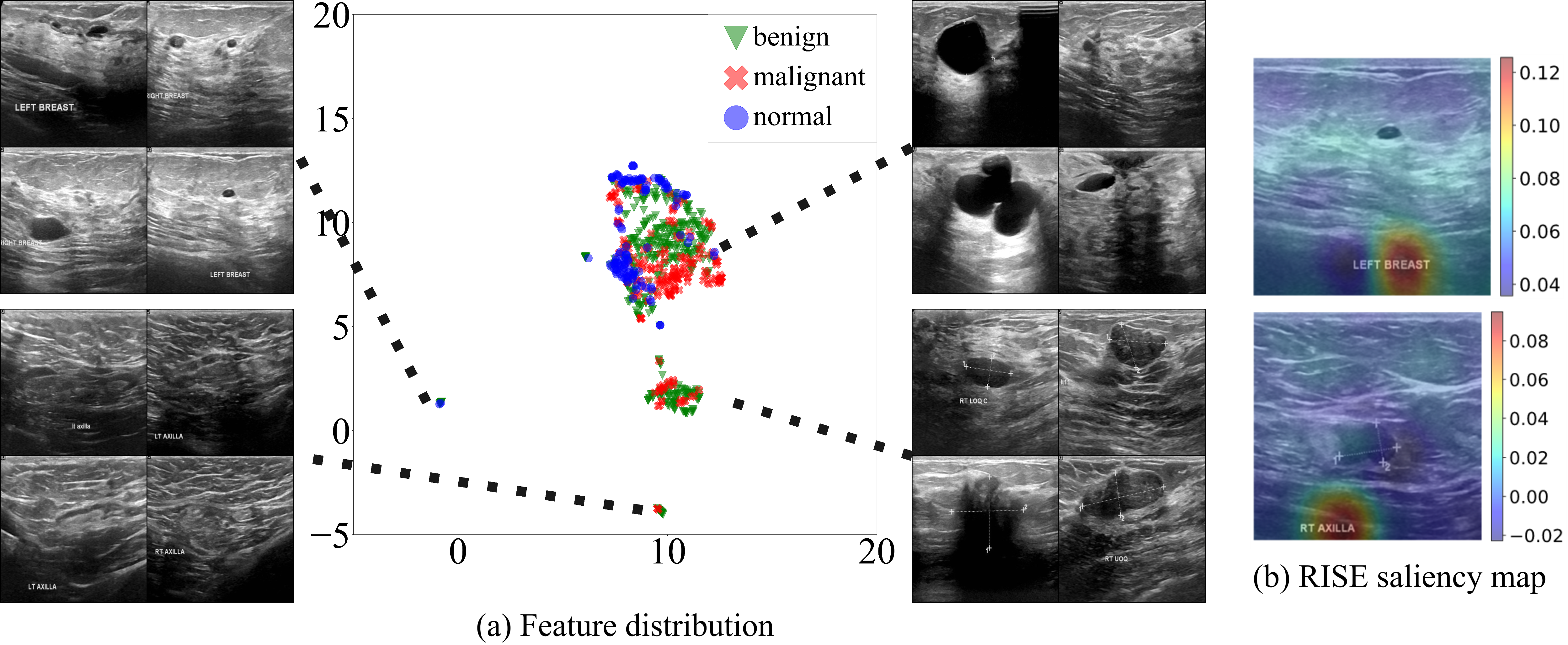}
    \caption{
    A example of feature distribution on the breast US~\cite{AlDhabyani20} Dataset and corresponding RISE saliency maps.
    (a) Feature distribution and example images from the Breast Ultrasound Images Dataset~\cite{AlDhabyani20}, generated via LLaVA-Med. The upper-left and lower-left images are labeled as 'BREAST' and 'AXILLA,' respectively. The tumours in the lower-right images are annotated with lines. (b) Class activation map from the LLM2CLIP and linear SVM classifiers. The model focus on the textual bias written on images.}
    \label{fig:FD_breastUS_llavamed}
    \end{center}
\end{figure}

The feature distributions produced by the VLMs are strongly influenced by background biases present in the images. 
Figure~\ref{fig:FD_breastUS_llavamed} shows the feature distributions on Breast US dataset, and illustrates the effect of background biases.
Breast US dataset contains images with overlaid textual notes and graphical annotations. 
ALL VLMs form distinct clusters based on these non-diagnostic, background elements. 
Notably, the models extract features discriminative for the semantic meaning of the overlaid words. 
The two small clusters at the bottom of Figure~\ref{fig:FD_breastUS_llavamed} are segmented based on the meaning of the text, with clusters forming around anatomical terms like "Breast" and "Axilla" rather than simple directional words like "Left" or "Right". 
This background text effects are already reported as~\cite{goh2021multimodal} in natural image domain. 
The VLMs formed another cluster based on the presence of line annotations. 
This sensitivity to background biases are observed across all datasets where such elements are present.

To confirm that these biases are being used for classification, we visualized the attention of a linear classifier using RISE, as shown in Figure~\ref{fig:FD_breastUS_llavamed}(b). 
The attention maps clearly show that the classifier focuses on the words "Breast" and "Axilla," confirming that background information are leveraged for discrimination. 
While this demonstrates the VLM's capability to capture a wide range of visual information, it also highlights a significant risk: if not properly addressed, models may rely on these spurious correlations during inference. 
Importantly, no significant difference about the effect of background bias are observed between medical-specialized and non-medical VLMs.

\subsection{Medical specialization doesn't specialize image encoders}\label{s:score}
\begin{figure}[tb]
    \begin{center}
    \includegraphics[width=\linewidth]{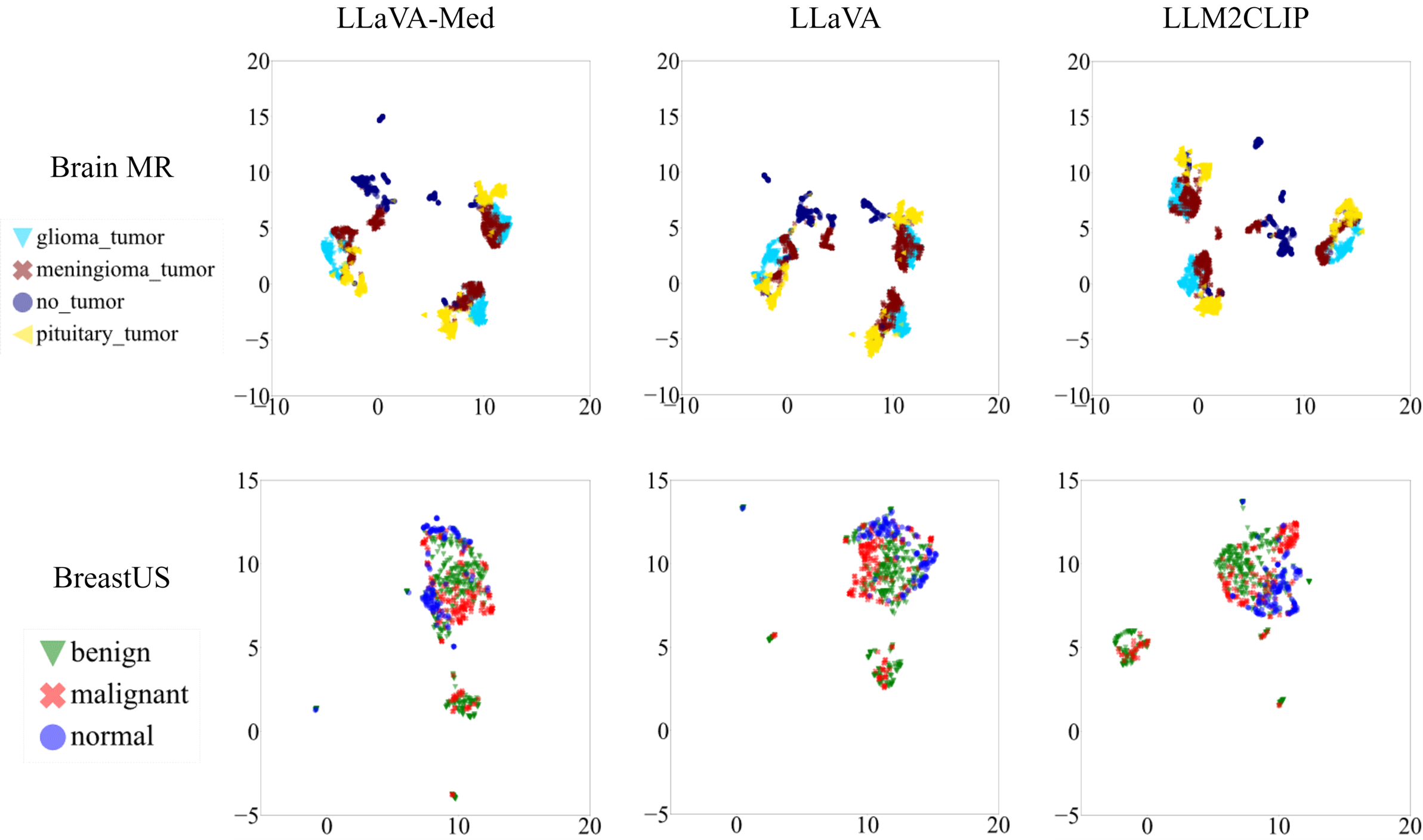}
    \caption{Examples of feature distributions for LLaVA-Med, its non-medical counterpart LLaVA, and LLM2CLIP employing an LLM as its text encoder. The upper row shows distributions in Brain MR~\cite{Bhuvaji20}, while the lower row shows distributions in Breast US~\cite{AlDhabyani20}. No significant differences are observed in the feature distributions between LLaVA-Med and other non-medical VLMs employing an LLM as their text encoder.}
    \label{fig:FD_noSpecializationEffect}
    \end{center}
\end{figure}

In our experiments, a qualitative analysis of the feature distributions did not reveal a clear advantage for models specialized in medical images. 
Figure~\ref{fig:FD_noSpecializationEffect} shows examples of feature distribution via medical and corresponding non-medical VLMs.
For instance, the distribution from LLaVA-Med shows no significant qualitative difference when compared to others from its non-medical counterparts. 
This suggests that, at least in terms of the visualized feature space structure, specialization on medical data alone does not guarantee a more organized or discriminative representation. 

The quantitative evaluation, presented in Tables~\ref{tab:classification_scores} and \ref{tab:classification_scores_2}, provides further insights into the discriminative capabilities of the high-dimensional feature spaces. 
Models employing an LLM as their text encoder, such as LLM2CLIP and LLaVA, achieved remarkable classification scores despite lacking pre-training on medical data. 
Notably, on the ISIC~2019 dataset, LLM2CLIP outperformed MONET, a dermatology-specific VLM.
Furthermore, LLM2CLIP demonstrated accuracy comparable to FLAIR, a VLM explicitly pre-trained on the DeepDRiD dataset. 
In the pathology domain, UNI, a foundation model trained in a self-supervised manner on one billion image patches, surpassed nearly all other models. 
These results suggest that enhancing the text encoder with an LLM may be more impactful for feature discrimination than simply amassing tens of millions of medical images for pre-training.

The strong performance of LLM2CLIP and other VLMs employing LLM can be attributed to its enhanced text encoder, which implies that previous models may not have fully leveraged the rich information contained in textual descriptions. 
Indeed, models incorporating LLMs, such as LLaVA and LLaVA-Med, consistently achieve high scores. 
Moreover, MedSigLIP, a modality-agnostic model that also uses an LLM, demonstrates even better performance. 
Therefore, a crucial direction for improving image feature extraction in medical VLMs is to enhance the text encoder, followed by re-training on medical data to fine-tune the representations.

\begin{table*}[tb]
    \centering
    \caption{Classification test accuracy and macro F1-score for models across datasets. "$^\dagger$" indicates the modality is included in training set. "*" indicates the dataset is included in training dataset of the model.}
    \label{tab:classification_scores}
    \hspace*{-2.5cm}
    \begin{tabular}{lcccccccc}
        \toprule
        \multirow{4}{*}{model name} 
        & \multicolumn{2}{c}{Brain MR~\cite{Bhuvaji20}} 
        & \multicolumn{2}{c}{SARS-COV-2 CTscan~\cite{soares20}} 
        & \multicolumn{2}{c}{PneumoniaMNIST~\cite{medmnistv1,medmnistv2,Kermany18}} 
        & \multicolumn{2}{c}{BreastUS~\cite{AlDhabyani20}} \\
        & $n=2,870$ & $2$ classes 
        & $n=2,481$ & $2$ classes 
        & $n=4,708$ & $14$ classes 
        & $n=780$ & $3$ classes \\
        \cmidrule(lr){2-3} \cmidrule(lr){4-5} \cmidrule(lr){6-7}  \cmidrule(lr){8-9}
        & Accuracy & F1-score & Accuracy & F1-score & Accuracy & F1-score & Accuracy & F1-score \\
        \midrule
        Medical VLMs                          &  &  &  &  &  \\
        MedSigLIP
        & $\boldsymbol{97\pm 0.5}\%^\dagger$ & $\boldsymbol{97\pm 0.5}\%^\dagger$ & $95\pm 0.2\%^\dagger$ & $95\pm 0.2\%^\dagger$ & $\boldsymbol{98\pm 0.3}\%^\dagger$ & $\boldsymbol{97\pm 0.3}\%^\dagger$ & $\boldsymbol{82\pm 2.3}\%^\dagger$ & $79\pm 2.3\%^\dagger$ \\
        BiomedCLIP~\cite{Zhang23}
        & $91\pm 0.5\%^\dagger$ & $91\pm 0.5\%^\dagger$ & $92\pm 0.6\%^\dagger$ & $92\pm 0.6\%^\dagger$ & $97\pm 0.1\%^\dagger$ & $96\pm 0.1\%^\dagger$ & $74\pm 2.0\%^\dagger$ & $71\pm 2.0\%^\dagger$ \\
        ConceptCLIP~\cite{Xie24} 
        & $95\pm 0.7\%^\dagger$ & $95\pm 0.7\%^\dagger$ & $94\pm 0.5\%^\dagger$ & $94\pm 0.5\%^\dagger$ & $97\pm 0.3\%^\dagger$ & $96\pm 0.3\%^\dagger$ & $78\pm 1.5\%^\dagger$ & $77\pm 1.5\%^\dagger$ \\
        LLaVA-Med~\cite{Li23} 
        & $93\pm 0.2\%^\dagger$ & $93\pm 0.2\%^\dagger$ & $95\pm 0.4\%^\dagger$ & $95\pm 0.4\%^\dagger$ & $96\pm 0.2\%^\dagger$ & $95\pm 0.2\%^\dagger$ & $\boldsymbol{82\pm 1.5}\%^\dagger$ & $\boldsymbol{80\pm 1.5}\%^\dagger$ \\
        LLaVA-Med++~\cite{Xie24} 
        & $93\pm 0.5\%^\dagger$ & $93\pm 0.5\%^\dagger$ & $94\pm 0.1\%^\dagger$ & $94\pm 0.1\%^\dagger$ & $96\pm 0.5\%^\dagger$ & $95\pm 0.5\%^\dagger$ & $80\pm 1.4\%^\dagger$ & $78\pm 1.4\%^\dagger$ \\
        \midrule
        \multicolumn{9}{l}{Modality-specific VLMs or large-scale medical pre-trained foundation models} \\
        CXR-CLIP~\cite{you23} 
        & $85\pm 0.7\%$ & $85\pm 0.7\%$ & $76\pm 1.5\%$ & $76\pm 1.5\%$ & $92\pm 0.5\%^\dagger$ & $90\pm 0.5\%^\dagger$ & $70\pm 1.9\%$ & $65\pm 1.9\%$ \\
        CONCH~\cite{Lu24} 
        & $91\pm 1.1\%$ & $91\pm 1.1\%$ & $92\pm 0.3\%$ & $92\pm 0.3\%$ & $97\pm 0.2\%$ & $96\pm 0.2\%$ & $79\pm 1.4\%$ & $75\pm 1.4\%$ \\
        UNI~\cite{Chen24} 
        & $93\pm 0.3\%$ & $93\pm 0.3\%$ & $\boldsymbol{96\pm 0.1}\%$ & $\boldsymbol{96\pm 0.1}\%$ & $97\pm 0.2\%$ & $97\pm 0.2\%$ & $77\pm 2.7\%$ & $74\pm 2.7\%$ \\
        FLAIR~\cite{Rodriguez25} 
        & $88\pm 0.4\%$ & $88\pm 0.4\%$ & $91\pm 0.7\%$ & $91\pm 0.7\%$ & $95\pm 0.3\%$ & $94\pm 0.3\%$ & $73\pm 1.5\%$ & $67\pm 1.5\%$ \\
        MONET~\cite{Rodriguez25} 
        & $92\pm 0.4\%$ & $92\pm 0.4\%$ & $94\pm 0.5\%$ & $94\pm 0.5\%$ & $96\pm 0.4\%$ & $95\pm 0.4\%$ & $77\pm 2.1\%$ & $73\pm 2.1\%$ \\
        \midrule
        \multicolumn{9}{l}{non-medical VLMs} \\
        CLIP (B/16)~\cite{Radford21} 
        & $91\pm 0.2\%$ & $91\pm 0.2\%$ & $90\pm 0.8\%$ & $90\pm 0.8\%$ & $96\pm 0.4\%$ & $95\pm 0.4\%$ & $78\pm 2.3\%$ & $75\pm 2.3\%$ \\
        CLIP (L/14)~\cite{Radford21} 
        & $90\pm 0.9\%$ & $90\pm 0.9\%$ & $89\pm 0.6\%$ & $89\pm 0.6\%$ & $96\pm 0.4\%$ & $95\pm 0.4\%$ & $77\pm 1.0\%$ & $74\pm 1.0\%$ \\
        CLIP (G/14)~\cite{Radford21} 
        & $89\pm 0.8\%$ & $89\pm 0.8\%$ & $92\pm 0.5\%$ & $92\pm 0.5\%$ & $97\pm 0.4\%$ & $96\pm 0.4\%$ & $81\pm 1.9\%$ & $79\pm 1.9\%$ \\
        EVA02~\cite{Fang24} 
        & $93\pm 0.4\%$ & $93\pm 0.4\%$ & $93\pm 0.5\%$ & $93\pm 0.5\%$ & $98\pm 0.4\%$ & $97\pm 0.4\%$ & $78\pm 1.4\%$ & $74\pm 1.4\%$ \\
        SigLIP~\cite{Xie24} 
        & $95\pm 0.7\%$ & $95\pm 0.7\%$ & $90\pm 0.6\%$ & $90\pm 0.6\%$ & $95\pm 0.4\%$ & $94\pm 0.4\%$ & $78\pm 1.8\%$ & $75\pm 1.8\%$ \\
        LLaVA~\cite{Liu23,Liu24}
        & $92\pm 0.5\%$ & $92\pm 0.5\%$ & $93\pm 0.4\%$ & $93\pm 0.4\%$ & $96\pm 0.1\%$ & $95\pm 0.1\%$ & $78\pm 1.1\%$ & $75\pm 1.1\%$ \\
        LLM2CLIP~\cite{Huang24} 
        & $94\pm 0.6\%$ & $94\pm 0.6\%$ & $93\pm 0.4\%$ & $93\pm 0.4\%$ & $97\pm 0.4\%$ & $96\pm 0.4\%$ & $\boldsymbol{82\pm 0.9}\%$ & $79\pm 0.9\%$ \\
        \midrule
        non VLMs &  &  &  &  &  & \\
        VGG16~\cite{SimonyanZ14}
        & $81\pm 0.6\%$ & $81\pm 0.6\%$ & $90\pm 0.7\%$ & $90\pm 0.7\%$ & $94\pm 0.3\%$ & $92\pm 0.3\%$ & $73\pm 1.3\%$ & $69\pm 1.3\%$ \\
        ResNet50~\cite{He16} 
        & $88\pm 1.0\%$ & $88\pm 1.0\%$ & $94\pm 0.3\%$ & $94\pm 0.3\%$ & $97\pm 0.3\%$ & $96\pm 0.3\%$ & $73\pm 2.0\%$ & $69\pm 2.0\%$ \\
        ViT-L-16~\cite{dosovitskiy21} 
        & $92\pm 0.6\%$ & $92\pm 0.6\%$ & $90\pm 0.6\%$ & $90\pm 0.6\%$ & $96\pm 0.5\%$ & $95\pm 0.5\%$ & $73\pm 2.1\%$ & $69\pm 2.1\%$ \\
        DINOv2
        & $92\pm 0.5\%$ & $92\pm 0.5\%$ & $94\pm 0.6\%$ & $94\pm 0.6\%$ & $97\pm 0.4\%$ & $96\pm 0.4\%$ & $76\pm 0.6\%$ & $73\pm 0.6\%$ \\
        \bottomrule \\
    \end{tabular}
\end{table*}

\begin{table*}[tb]
    \centering
    \caption{Classification test scores for the other modalities than Table~\ref{tab:classification_scores}. }
    \label{tab:classification_scores_2}
    \hspace*{-2.6cm}
    \begin{tabular}{lcccccccc}
        \toprule
        \multirow{3}{*}{model name} 
        & \multicolumn{2}{c}{BreakHis~\cite{Fabio16}} 
        & \multicolumn{2}{c}{HiCervix~\cite{Cai24}} 
        & \multicolumn{2}{c}{DeepDRiD~\cite{Liu22}} 
        & \multicolumn{2}{c}{ISIC~2019~\cite{Tschandl18,Codella18,Hernandez24}} \\
        & $n=7,909$ & $2$ classes 
        & $n=28,160$ & $29$ classes 
        & $n=1,200$ & $5$ classes 
        & $n=25,331$ & $9$ classes \\
        \cmidrule(lr){2-3} \cmidrule(lr){4-5} \cmidrule(lr){6-7}  \cmidrule(lr){8-9}
        & Accuracy & F1-score & Accuracy & F1-score & Accuracy & F1-score & Accuracy & F1-score \\
        \midrule
        Medical VLMs                          &  &  &  &  &  \\
        MedSigLIP
        & $87\pm 1.4\%^\dagger$ & $90\pm 1.4\%^\dagger$ & $\boldsymbol{64\pm 0.2}\%^\dagger$ & $\boldsymbol{60\pm 0.2}\%^\dagger$ & $66\pm 2.0\%^\dagger$ & $56\pm 2.0\%^\dagger$ & $\boldsymbol{77\pm 0.2}\%^\dagger$ & $\boldsymbol{65\pm 0.2}\%^\dagger$ \\
        BiomedCLIP~\cite{Zhang23}
        & $86\pm 0.9\%^\dagger$ & $89\pm 0.9\%^\dagger$ & $58\pm 0.3\%^\dagger$ & $54\pm 0.3\%^\dagger$ & $55\pm 1.1\%^\dagger$ & $42\pm 1.1\%^\dagger$ & $71\pm 0.1\%^\dagger$ & $54\pm 0.1\%^\dagger$ \\
        ConceptCLIP~\cite{Xie24} 
        & $86\pm 0.4\%^\dagger$ & $89\pm 0.4\%^\dagger$ & $61\pm 0.1\%^\dagger$ & $58\pm 0.1\%^\dagger$ & $55\pm 1.6\%^\dagger$ & $42\pm 1.6\%^\dagger$ & $75\pm 0.3\%^\dagger$ & $61\pm 0.3\%^\dagger$ \\
        LLaVA-Med~\cite{Li23} 
        & $86\pm 0.9\%^\dagger$ & $89\pm 0.9\%^\dagger$ & $54\pm 0.4\%^\dagger$ & $51\pm 0.4\%^\dagger$ & $57\pm 1.5\%^\dagger$ & $48\pm 1.5\%^\dagger$ & $73\pm 0.2\%^\dagger$ & $58\pm 0.2\%^\dagger$ \\
        LLaVA-Med++~\cite{Xie24} 
        & $85\pm 1.3\%^\dagger$ & $88\pm 1.3\%^\dagger$ & $54\pm 0.5\%^\dagger$ & $52\pm 0.5\%^\dagger$ & $57\pm 1.0\%^\dagger$ & $47\pm 1.0\%^\dagger$ & $73\pm 0.4\%^\dagger$ & $57\pm 0.4\%^\dagger$ \\
        \midrule
        \multicolumn{9}{l}{Modality-specific VLMs or large-scale medical pre-trained foundation models} \\
        CXR-CLIP~\cite{you23} 
        & $81\pm 0.6\%$ & $86\pm 0.6\%$ & $46\pm 0.4\%$ & $41\pm 0.4\%$ & $50\pm 2.4\%$ & $35\pm 2.4\%$ & $61\pm 0.3\%$ & $33\pm 0.3\%$ \\
        CONCH~\cite{Lu24} 
        & $89\pm 0.6\%^\dagger$ & $91\pm 0.6\%^\dagger$ & $59\pm 0.4\%$ & $55\pm 0.4\%$ & $55\pm 3.0\%$ & $45\pm 3.0\%$ & $72\pm 0.2\%$ & $55\pm 0.2\%$ \\
        UNI~\cite{Chen24} 
        & $\boldsymbol{91\pm 0.9}\%^\dagger$ & $\boldsymbol{93\pm 0.9}\%^\dagger$ & $62\pm 0.3\%$ & $59\pm 0.3\%$ & $57\pm 1.6\%$ & $46\pm 1.6\%$ & $74\pm 0.2\%$ & $61\pm 0.2\%$ \\
        FLAIR~\cite{Rodriguez25} 
        & $83\pm 1.1\%$ & $88\pm 1.1\%$ & $53\pm 0.3\%$ & $48\pm 0.3\%$ & $\boldsymbol{67\pm 2.0}\%^*$ & $\boldsymbol{61\pm 2.0}\%^*$ & $66\pm 0.3\%$ & $44\pm 0.3\%$ \\
        MONET~\cite{Rodriguez25} 
        & $85\pm 1.3\%$ & $88\pm 1.3\%$ & $59\pm 0.3\%$ & $55\pm 0.3\%$ & $52\pm 2.7\%$ & $41\pm 2.7\%$ & $74\pm 0.1\%^\dagger$ & $59\pm 0.1\%^\dagger$ \\
        \midrule
        \multicolumn{9}{l}{non-medical VLMs} \\
        CLIP (B/16)~\cite{Radford21} 
        & $84\pm 1.0\%$ & $87\pm 1.0\%$ & $57\pm 0.1\%$ & $54\pm 0.1\%$ & $51\pm 2.6\%$ & $41\pm 2.6\%$ & $71\pm 0.3\%$ & $51\pm 0.3\%$ \\
        CLIP (L/14)~\cite{Radford21} 
        & $84\pm 1.2\%$ & $87\pm 1.2\%$ & $58\pm 0.3\%$ & $54\pm 0.3\%$ & $50\pm 3.0\%$ & $39\pm 3.0\%$ & $73\pm 0.1\%$ & $56\pm 0.1\%$ \\
        CLIP (G/14)~\cite{Radford21} 
        & $84\pm 0.7\%$ & $88\pm 0.7\%$ & $58\pm 0.2\%$ & $54\pm 0.2\%$ & $55\pm 1.8\%$ & $42\pm 1.8\%$ & $74\pm 0.3\%$ & $60\pm 0.3\%$ \\
        EVA02~\cite{Fang24} 
        & $85\pm 0.4\%$ & $88\pm 0.4\%$ & $60\pm 0.2\%$ & $56\pm 0.2\%$ & $52\pm 2.3\%$ & $39\pm 2.3\%$ & $75\pm 0.2\%$ & $61\pm 0.2\%$ \\
        SigLIP~\cite{Xie24} 
        & $85\pm 1.2\%$ & $88\pm 1.2\%$ & $59\pm 0.4\%$ & $56\pm 0.4\%$ & $53\pm 1.3\%$ & $42\pm 1.3\%$ & $76\pm 0.2\%$ & $61\pm 0.2\%$ \\
        LLaVA~\cite{Liu23,Liu24}
        & $86\pm 1.0\%$ & $89\pm 1.0\%$ & $56\pm 0.5\%$ & $53\pm 0.5\%$ & $57\pm 2.2\%$ & $49\pm 2.2\%$ & $74\pm 0.3\%$ & $61\pm 0.3\%$ \\
        LLM2CLIP~\cite{Huang24} 
        & $87\pm 0.7\%$ & $90\pm 0.7\%$ & $60\pm 0.4\%$ & $57\pm 0.4\%$ & $59\pm 1.4\%$ & $47\pm 1.4\%$ & $76\pm 0.3\%$ & $64\pm 0.3\%$ \\
        \midrule
        non VLMs &  &  &  &  &  & \\
        VGG16~\cite{SimonyanZ14}
        & $82\pm 0.6\%$ & $86\pm 0.6\%$ & $51\pm 0.2\%$ & $47\pm 0.2\%$ & $50\pm 0.6\%$ & $37\pm 0.6\%$ & $65\pm 0.2\%$ & $40\pm 0.2\%$ \\
        ResNet50~\cite{He16} 
        & $85\pm 0.9\%$ & $88\pm 0.9\%$ & $50\pm 0.3\%$ & $46\pm 0.3\%$ & $51\pm 0.9\%$ & $37\pm 0.9\%$ & $68\pm 0.2\%$ & $50\pm 0.2\%$ \\
        ViT-L-16~\cite{dosovitskiy21} 
        & $85\pm 1.5\%$ & $88\pm 1.5\%$ & $58\pm 0.4\%$ & $54\pm 0.4\%$ & $53\pm 1.2\%$ & $42\pm 1.2\%$ & $73\pm 0.4\%$ & $59\pm 0.4\%$ \\
        DINOv2
        & $85\pm 0.9\%$ & $88\pm 0.9\%$ & $58\pm 0.2\%$ & $54\pm 0.2\%$ & $59\pm 2.5\%$ & $51\pm 2.5\%$ & $\boldsymbol{77\pm 0.1}\%$ & $\boldsymbol{65\pm 0.1}\%$ \\
        \bottomrule
    \end{tabular}
\end{table*}

\subsection{Negative effect of pre-training on PMC-15M}
While our analysis of feature distributions did not consistently show an advantage for medical-specialized models, we identified instances where such specialization led to inferior feature extraction. 
Figure~\ref{fig:FD_noSpecializationEffect} illustrates this by visualizing the feature distributions for microbe subclasses in cervical cytology images. 
Compared to the three distributions on the left, those on the right exhibit a more discriminative structure, forming distinct clusters for each class. 
The models on the left are modality-agnostic VLMs primarily trained on the PMC-15M dataset, which is sourced from medical papers and likely contains some cervical cytology images. 
In contrast, the models on the right, a non-medical VLM (top) and a modality-specific model trained on a different modality (bottom), are not specialized on cervical cytology images. 
This leads to the counter-intuitive finding that features from models trained for general medical use can be less discriminative than those from non-medical VLMs for this task.

We attribute this nature to the quality of the publication-derived data. 
As reported by Zhang~\textit{et al.}, PMC-15M contains a significant number of non-medical images, such as statistical figures, flowcharts, and mathematical formulas. 
While training on such noisy data may enable a model to distinguish between medical and non-medical images, it can degrade its ability to extract fine-grained features from high-quality, clean data. 
The cervical cytology images used in this study are of high quality, lacking background biases like overlaid text. 
For such data, non-medical VLMs trained on massive, curated datasets or modality-specific models trained on high-quality data may extract more discriminative features. 
Although cervical cytology is a crucial task, it is not as common as major modalities like MRI or pathology. 
Consequently, for less common tasks involving high-quality images, leveraging a modality-agnostic medical VLM is not necessarily the optimal choice. 
As our results in Section~\ref{s:score} suggest, models that employ an LLM as the text encoder may represent a better alternative.

\begin{figure}[tb]
    \begin{center}
    \includegraphics[width=\linewidth]{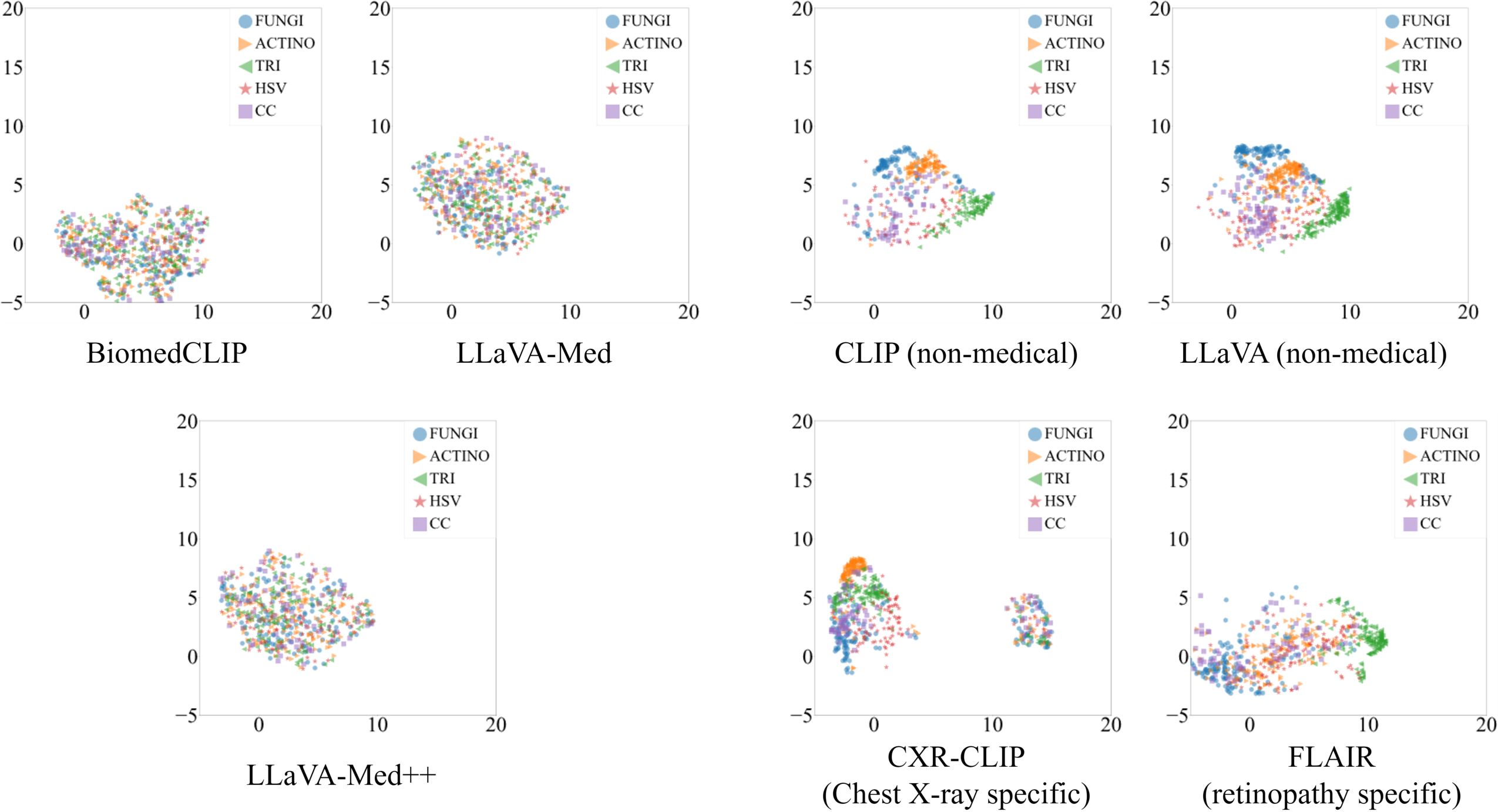}
    \caption{Feature distributions in microbe subclasses on HiCervix~\cite{Cai24} by LLMs trained on PMC-15M~\cite{Zhang23} and other VLMs. The three on the left (BiomedCLIP~\cite{Zhang23}, LLaVA-Med~\cite{Li23}, LLaVA-Med++~\cite{Xie24}) are feature distributions from modality-agnostic VLMs trained on PMC-15M, which can be assumed to include cytological images. The two on the upper right (CLIP~\cite{Radford21}, LLaVA~\cite{Liu23, Liu24}) are non-medical VLMs. The CXR-CLIP~\cite{you23} and FLAIR~\cite{silva2025foundation} on the lower right are VLMs specialised for modalities other than cytology, and cytological images were not included in their training data. The model trained on PMC-15M shows no disentangled cluster within the subclasses. Conversely, the VLM on the right, which did not learn cervical images, exhibits a clear cluster structure for each subclass.}
    \label{fig:FD_PMC15M}
    \end{center}
\end{figure}

\section{Limitations}

While this paper has demonstrated the capabilities and weaknesses of existing medical VLMs, their superiority is assumed to vary for tasks beyond two-dimensional image classification.
For instance, our investigation has not extended to tasks involving the analysis of 3D volume data, the utilisation of WSI on pathology, or multi-modal image analysis.
As some VLMs for volume data~\cite{Hamamci24} and WSI analysis~\cite{Chen24} have been emerged, these models would clearly hold an advantage over existing models.

This study focuses exclusively on the analysis of image-level features. However, the primary contribution of VLMs lies in the integration of visual and linguistic representations within a embedding space. 
Modality-specific models, in particular, are expected to learn domain-specific linguistic patterns and clinical terminologies that frequently occur within their respective fields. 
Such specialization likely enhances performance in zero-shot classification tasks by leveraging the capabilities of the text encoder.
While the present work provides insights into visual feature distributions, a comprehensive investigation into text-based features and their alignment with images remains a critical direction for future research. 
Further exploration of how textual cues influence the formation of the universal feature space will be essential to fully elucidating the potential of medical VLMs.

Local image representations are also out of scope of this study. 
In medical image processing, there is often significant interest in feature representations within local regions of interest, such as tumour areas. 
Indeed, numerous medical segmentation foundation models have been derived to meet this demand. 
By verifying local feature representations, we can expect to uncover new insights, such as eliminating the significant background bias demonstrated in this research and elucidating the relationship between anatomical structures and tumour features.

In this paper, there are a number of VLMs that were mentioned in Table.~\ref{tab:famous_VLMs}, but it is noted that some of these could not be experimented with. 
The primarily reason of this is the difficulties of prepairing exmperimental emvironment. 
To illustrate this point, consider the example of LLaVA: with the advent of LLaVA-NeXT~\cite{li2024llava}, a new class for LLaVA was defined. 
LLaVA for natural images could be loaded with warnings now, but llava-med, which is stored in Hugging~Face, generats errors when loaded and are unavailable (confirmed in version 4.57.1). 
This issue was traced back to a deficiency in the config file. 

Another limitation of this study is about preprocessing of medical images.
In our experiments, we utilized the default pre-processing functions provided with each official implementation of VLM.
These functions are typically optimized for standard 8-bit RGB color images (e.g., PNG or JPEG formats).
However, in clinical practice, medical images are often consists in DICOM format and consists of grayscale images with a high dynamic range (e.g., 12-bit or 16-bit depth), which differs significantly from the distributions of natural images.
While specialized pre-processing is essential to handle the unique value ranges and bit depths of medical modalities, the robustness of existing VLMs to these diverse data formats remains largely unverified. 
Since feature extraction is highly sensitive to initial transformations such as normalization and bit-depth rescaling, the choice of pre-processing likely exerts a substantial influence on the resulting latent representations. 
Consequently, identifying the optimal pre-processing strategies to maximize the performance of medical VLMs for specific downstream tasks remains an important direction for future research.

This predicament carries significant ramifications, including the protracted development of medical VLMs, even within substantial teams, and the uncertainty surrounding their sustained utilisation. 
In particular, the transformers~\cite{wolf-etal-2020-transformers}, that is to say the Hugging Face readout library and the main repository for many VLMs, is frequently specified and modified to suit the latest models. 
This has resulted in considerable challenges in terms of reading out medical VLMs through conventional means. 
This is primarily due to the fact that specification changes are primarily focused on the latest non-medical VLMs, which often results in medical VLMs being unable to adapt to these new specifications and being unable to be read out.
Of course, it is possible to make them work by adjusting the library version, but this requires a lot of effort. 
The preparation of a distinct environment for each comparison is expensive, even if a virtual environment such as Docker is used, and it can be difficult to unify the experimental environment due to differences in library versions. 
Such difficulties in building a development environment are considered to be one of the reasons for the slow development of medical VLM.

\section{Conclusion}
This study investigated the feature distributions of medical and non-medical VLMs to assess the true impact of medical specialization. Our analysis across eight diverse medical imaging datasets revealed a key insight: the benefits of specializing a VLM's image encoder on medical data are not as clear-cut as expected.

A principal finding is that enhancing text encoder is often more crucial than the image encoder's medical pre-training. We found that non-medical VLMs with enhanced text encoders produced highly refined features. Consequently, their classification performance was competitive with, and sometimes superior to, specialized medical models.

Furthermore, our experiments uncovered a significant vulnerability in both medical and non-medical VLMs: a strong susceptibility to background bias. We found that models are easily fooled by spurious information, such as overlaid text annotations. Attention maps confirmed that classifiers often rely on this non-diagnostic data, posing a significant risk for clinical deployment.

We also observed a counter-intuitive result: some medical VLMs produced less discriminative features than non-medical models when analyzing clean images. This was especially true for models trained on large, noisy, publication-derived datasets like PMC-15M. This suggests that poor-quality pre-training data can degrade a model's ability to extract fine-grained features.

These findings suggest a strategic shift for developing medical VLMs. Future efforts may benefit more from enhancing the text encoder's contextual understanding (e.g., using LLMs) rather than pre-training image encoders on massive, noisy datasets. A more effective approach may be to combine a strong text encoder with fine-tuning on high-quality, curated medical data. Future work should extend this analysis to text features, tasks beyond 2D classification (like 3D or WSI analysis), and local feature representations.

\bibliographystyle{unsrt}  
\bibliography{mybib}

@inproceedings{petsiuk2018rise,
  title = {RISE: Randomized Input Sampling for Explanation of Black-box Models},
  author = {Vitali Petsiuk and Abir Das and Kate Saenko},
  booktitle = {Proceedings of the British Machine Vision Conference (BMVC)},
  year = {2018}
}

@article{Zhang24,
  title={Vision-language models for vision tasks: A survey},
  author={Zhang, Jingyi and Huang, Jiaxing and Jin, Sheng and Lu, Shijian},
  journal={IEEE Transactions on Pattern Analysis and Machine Intelligence},
  year={2024},
  publisher={IEEE}
}

@article{Shrestha23,
  title={Medical Vision Language Pretraining: A survey},
  author={Shrestha, Prashant and Amgain, Sanskar and Khanal, Bidur and Linte, Cristian A and Bhattarai, Binod},
  journal={arXiv preprint arXiv:2312.06224},
  year={2023}
}

@inproceedings{Radford21,
  title={Learning transferable visual models from natural language supervision},
  author={Radford, Alec and Kim, Jong Wook and Hallacy, Chris and Ramesh, Aditya and Goh, Gabriel and Agarwal, Sandhini and Sastry, Girish and Askell, Amanda and Mishkin, Pamela and Clark, Jack and others},
  booktitle={International conference on machine learning},
  pages={8748--8763},
  year={2021},
  organization={PMLR}
}

@inproceedings{Liu23,
  author = {Liu, Haotian and Li, Chunyuan and Wu, Qingyang and Lee, Yong Jae},
  booktitle = {Advances in Neural Information Processing Systems},
  title = {Visual Instruction Tuning},
  year = {2023}
}

@InProceedings{Liu24,
  author    = {Liu, Haotian and Li, Chunyuan and Li, Yuheng and Lee, Yong Jae},
  title     = {Improved Baselines with Visual Instruction Tuning},
  booktitle = {Proceedings of the IEEE/CVF Conference on Computer Vision and Pattern Recognition (CVPR)},
  month     = {June},
  year      = {2024},
  pages     = {26296-26306}
}

@article{blankemeier2024merlin,
  title={Merlin: A vision language foundation model for 3d computed tomography},
  author={Blankemeier, Louis and Cohen, Joseph Paul and Kumar, Ashwin and Van Veen, Dave and Gardezi, Syed Jamal Safdar and Paschali, Magdalini and Chen, Zhihong and Delbrouck, Jean-Benoit and Reis, Eduardo and Truyts, Cesar and others},
  journal={Research Square},
  pages={rs--3},
  year={2024}
}

@inproceedings{wu2024mm,
  title={MM-retinal: Knowledge-enhanced foundational pretraining with fundus image-text expertise},
  author={Wu, Ruiqi and Zhang, Chenran and Zhang, Jianle and Zhou, Yi and Zhou, Tao and Fu, Huazhu},
  booktitle={International Conference on Medical Image Computing and Computer-Assisted Intervention},
  pages={722--732},
  year={2024},
  organization={Springer}
}

@article{nie2025explainable,
  title={An Explainable Biomedical Foundation Model via Large-Scale Concept-Enhanced Vision-Language Pre-training},
  author={Nie, Yuxiang and He, Sunan and Bie, Yequan and Wang, Yihui and Chen, Zhixuan and Yang, Shu and Cai, Zhiyuan and Wang, Hongmei and Wang, Xi and Luo, Luyang and others},
  journal={arXiv preprint arXiv:2501.15579},
  year={2025}
}

@misc{Eyepack,
    author = {Emma Dugas and Jared and Jorge and Will Cukierski},
    title = {Diabetic Retinopathy Detection},
    year = {2015},
    howpublished = {\url{https://kaggle.com/competitions/diabetic-retinopathy-detection}},
    note = {Kaggle}
}

@ARTICLE{AIROGS,
  author={de Vente, Coen and Vermeer, Koenraad A. and Jaccard, Nicolas and Wang, He and Sun, Hongyi and Khader, Firas and Truhn, Daniel and Aimyshev, Temirgali and Zhanibekuly, Yerkebulan and Le, Tien-Dung and Galdran, Adrian and Ballester, Miguel Ángel González and Carneiro, Gustavo and Devika, R. G. and Sethumadhavan, Hrishikesh Panikkasseril and Puthussery, Densen and Liu, Hong and Yang, Zekang and Kondo, Satoshi and Kasai, Satoshi and Wang, Edward and Durvasula, Ashritha and Heras, Jónathan and Zapata, Miguel Ángel and Araújo, Teresa and Aresta, Guilherme and Bogunović, Hrvoje and Arikan, Mustafa and Lee, Yeong Chan and Cho, Hyun Bin and Choi, Yoon Ho and Qayyum, Abdul and Razzak, Imran and van Ginneken, Bram and Lemij, Hans G. and Sánchez, Clara I.},
  journal={IEEE Transactions on Medical Imaging}, 
  title={AIROGS: Artificial Intelligence for Robust Glaucoma Screening Challenge}, 
  year={2024},
  volume={43},
  number={1},
  pages={542-557},
  doi={10.1109/TMI.2023.3313786}
}

@article{kim2024transparent,
    title = {Transparent medical image AI via an image-text foundation model grounded in medical literature},
    author = {Kim, Chanwoo and Gadgil, Soham U. and {DeGrave}, Alex J. and Omiye, Jesutofunmi A. and Cai, Zhuo Ran and Daneshjou, Roxana and Lee, Su-In},
    journal={Nature Medicine},
    volume = {30},
    number = {4},
    year={2024},
    pages = {1154--1165},
    doi={10.1038/s41591-024-02887-x},
    issn = {1546-170X},
    url={https://doi.org/10.1038/s41591-024-02887-x}
}

@InProceedings{Yan25make,
        author = { Yan, Siyuan AND Li, Xieji AND Hu, Ming AND Jiang, Yiwen AND Yu, Zhen AND Ge, Zongyuan},
        title = { { MAKE: Multi-Aspect Knowledge-Enhanced Vision-Language Pretraining for Zero-shot Dermatological Assessment } },
        booktitle = {proceedings of Medical Image Computing and Computer Assisted Intervention -- MICCAI 2025},
        year = {2025},
        publisher = {Springer Nature Switzerland},
        volume = {LNCS 15964},
        month = {September},
        page = {369 -- 379}
}

@inproceedings{irvin2019chexpert,
  title={Chexpert: A large chest radiograph dataset with uncertainty labels and expert comparison},
  author={Irvin, Jeremy and Rajpurkar, Pranav and Ko, Michael and Yu, Yifan and Ciurea-Ilcus, Silviana and Chute, Chris and Marklund, Henrik and Haghgoo, Behzad and Ball, Robyn and Shpanskaya, Katie and others},
  booktitle={Proceedings of the AAAI conference on artificial intelligence},
  volume={33},
  number={01},
  pages={590--597},
  year={2019}
}

@article{johnson2024mimic,
  title={Mimic-cxr database},
  author={Johnson, Alistair and Pollard, Tom and Mark, Roger and Berkowitz, Seth and Horng, Steven},
  journal={PhysioNet10},
  volume={13026},
  pages={C2JT1Q},
  year={2024}
}

@inproceedings{ghosh2024mammo,
  title={Mammo-clip: A vision language foundation model to enhance data efficiency and robustness in mammography},
  author={Ghosh, Shantanu and Poynton, Clare B and Visweswaran, Shyam and Batmanghelich, Kayhan},
  booktitle={International conference on medical image computing and computer-assisted intervention},
  pages={632--642},
  year={2024},
  organization={Springer}
}

@inproceedings{tan2019efficientnet,
  title={Efficientnet: Rethinking model scaling for convolutional neural networks},
  author={Tan, Mingxing and Le, Quoc},
  booktitle={International conference on machine learning},
  pages={6105--6114},
  year={2019},
  organization={PMLR}
}

@article{yan2025multimodal,
  title={A multimodal vision foundation model for clinical dermatology},
  author={Yan, Siyuan and Yu, Zhen and Primiero, Clare and Vico-Alonso, Cristina and Wang, Zhonghua and Yang, Litao and Tschandl, Philipp and Hu, Ming and Ju, Lie and Tan, Gin and others},
  journal={Nature Medicine},
  pages={1--12},
  year={2025},
  publisher={Nature Publishing Group}
}

@InProceedings{Shi_2024_CVPR,
    author    = {Shi, Jiangbo and Li, Chen and Gong, Tieliang and Zheng, Yefeng and Fu, Huazhu},
    title     = {ViLa-MIL: Dual-scale Vision-Language Multiple Instance Learning for Whole Slide Image Classification},
    booktitle = {Proceedings of the IEEE/CVF Conference on Computer Vision and Pattern Recognition (CVPR)},
    month     = {June},
    year      = {2024},
    pages     = {11248-11258}
}

@article{Zhang23,
  title={BiomedCLIP: a multimodal biomedical foundation model pretrained from fifteen million scientific image-text pairs},
  author={Zhang, Sheng and Xu, Yanbo and Usuyama, Naoto and Xu, Hanwen and Bagga, Jaspreet and Tinn, Robert and Preston, Sam and Rao, Rajesh and Wei, Mu and Valluri, Naveen and others},
  journal={arXiv preprint arXiv:2303.00915},
  year={2023}
}

@inproceedings{Li23,
  author = {Li, Chunyuan and Wong, Cliff and Zhang, Sheng and Usuyama, Naoto and Liu, Haotian and Yang, Jianwei and Naumann, Tristan and Poon, Hoifung and Gao, Jianfeng},
  booktitle = {Advances in Neural Information Processing Systems},
  title = {LLaVA-Med: Training a Large Language-and-Vision Assistant for Biomedicine in One Day},
  year = {2023}
}

@inproceedings{dosovitskiy21,
title={An Image is Worth 16x16 Words: Transformers for Image Recognition at Scale},
author={Alexey Dosovitskiy and Lucas Beyer and Alexander Kolesnikov and Dirk Weissenborn and Xiaohua Zhai and Thomas Unterthiner and Mostafa Dehghani and Matthias Minderer and Georg Heigold and Sylvain Gelly and Jakob Uszkoreit and Neil Houlsby},
booktitle={International Conference on Learning Representations},
year={2021},
url={https://openreview.net/forum?id=YicbFdNTTy}
}

@InProceedings{Thrush22,
    author    = {Thrush, Tristan and Jiang, Ryan and Bartolo, Max and Singh, Amanpreet and Williams, Adina and Kiela, Douwe and Ross, Candace},
    title     = {Winoground: Probing Vision and Language Models for Visio-Linguistic Compositionality},
    booktitle = {Proceedings of the IEEE/CVF Conference on Computer Vision and Pattern Recognition (CVPR)},
    month     = {June},
    year      = {2022},
    pages     = {5238-5248}
}

@inproceedings{Tu23,
  author = {Tu, Weijie and Deng, Weijian and Gedeon, Tom},
  booktitle = {Advances in Neural Information Processing Systems},
  title = {A Closer Look at the Robustness of Contrastive Language-Image Pre-Training (CLIP)},
  year = {2023}
}

@inproceedings{He16,
  title={Deep residual learning for image recognition},
  author={He, Kaiming and Zhang, Xiangyu and Ren, Shaoqing and Sun, Jian},
  booktitle={Proceedings of the IEEE conference on computer vision and pattern recognition},
  pages={770--778},
  year={2016}
}

@article{chen2020simple,
  title={A Simple Framework for Contrastive Learning of Visual Representations},
  author={Chen, Ting and Kornblith, Simon and Norouzi, Mohammad and Hinton, Geoffrey},
  journal={arXiv preprint arXiv:2002.05709},
  year={2020}
}

@article{chen2020big,
  title={Big Self-Supervised Models are Strong Semi-Supervised Learners},
  author={Chen, Ting and Kornblith, Simon and Swersky, Kevin and Norouzi, Mohammad and Hinton, Geoffrey},
  journal={arXiv preprint arXiv:2006.10029},
  year={2020}
}

@inproceedings{zhang2022contrastive,
  title={Contrastive learning of medical visual representations from paired images and text},
  author={Zhang, Yuhao and Jiang, Hang and Miura, Yasuhide and Manning, Christopher D and Langlotz, Curtis P},
  booktitle={Machine learning for healthcare conference},
  pages={2--25},
  year={2022},
  organization={PMLR}
}

@inproceedings{joulin2016learning,
  title={Learning visual features from large weakly supervised data},
  author={Joulin, Armand and Van Der Maaten, Laurens and Jabri, Allan and Vasilache, Nicolas},
  booktitle={European conference on computer vision},
  pages={67--84},
  year={2016},
  organization={Springer}
}

@misc{Leland20,
      title={UMAP: Uniform Manifold Approximation and Projection for Dimension Reduction}, 
      author={Leland McInnes and John Healy and James Melville},
      year={2020},
      eprint={1802.03426},
      archivePrefix={arXiv},
      primaryClass={stat.ML}
}

@misc{Huang24,
      title={LLM2CLIP: Powerful Language Model Unlock Richer Visual Representation}, 
      author={Weiquan Huang and Aoqi Wu and Yifan Yang and Xufang Luo and Yuqing Yang and Liang Hu and Qi Dai and Xiyang Dai and Dongdong Chen and Chong Luo and Lili Qiu},
      year={2024},
      eprint={2411.04997},
      archivePrefix={arXiv},
      primaryClass={cs.CV},
      url={https://arxiv.org/abs/2411.04997}, 
}

@article{Khattak24,
  title={UniMed-CLIP: Towards a Unified Image-Text Pretraining Paradigm for Diverse Medical Imaging Modalities},
  author={Khattak, Muhammad Uzair and Kunhimon, Shahina and Naseer, Muzammal and Khan, Salman and Khan, Fahad Shahbaz},
  journal={arXiv preprint arXiv:2412.10372},
  year={2024}
}

@misc{Xie24,
      title={MedTrinity-25M: A Large-scale Multimodal Dataset with Multigranular Annotations for Medicine}, 
      author={Yunfei Xie and Ce Zhou and Lang Gao and Juncheng Wu and Xianhang Li and Hong-Yu Zhou and Sheng Liu and Lei Xing and James Zou and Cihang Xie and Yuyin Zhou},
      year={2024},
      eprint={2408.02900},
      archivePrefix={arXiv},
      primaryClass={cs.CV},
      url={https://arxiv.org/abs/2408.02900}, 
}

@article{Lu24,
  title={A visual-language foundation model for computational pathology},
  author={Lu, Ming Y and Chen, Bowen and Williamson, Drew FK and Chen, Richard J and Liang, Ivy and Ding, Tong and Jaume, Guillaume and Odintsov, Igor and Le, Long Phi and Gerber, Georg and others},
  journal={Nature Medicine},
  pages={863-874},
  volume={30},
  year={2024},
  publisher={Nature Publishing Group}
}

@article{Chen24,
  title={Towards a General-Purpose Foundation Model for Computational Pathology},
  author={Chen, Richard J and Ding, Tong and Lu, Ming Y and Williamson, Drew FK and Jaume, Guillaume and Chen, Bowen and Zhang, Andrew and Shao, Daniel and Song, Andrew H and Shaban, Muhammad and others},
  journal={Nature Medicine},
  publisher={Nature Publishing Group},
  year={2024}
}

@Article{Wang24,
author={Wang, Xiyue
and Zhao, Junhan
and Marostica, Eliana
and Yuan, Wei
and Jin, Jietian
and Zhang, Jiayu
and Li, Ruijiang
and Tang, Hongping
and Wang, Kanran
and Li, Yu
and Wang, Fang
and Peng, Yulong
and Zhu, Junyou
and Zhang, Jing
and Jackson, Christopher R.
and Zhang, Jun
and Dillon, Deborah
and Lin, Nancy U.
and Sholl, Lynette
and Denize, Thomas
and Meredith, David
and Ligon, Keith L.
and Signoretti, Sabina
and Ogino, Shuji
and Golden, Jeffrey A.
and Nasrallah, MacLean P.
and Han, Xiao
and Yang, Sen
and Yu, Kun-Hsing},
title={A pathology foundation model for cancer diagnosis and prognosis prediction},
journal={Nature},
year={2024},
month={Oct},
day={01},
volume={634},
number={8035},
pages={970-978},
issn={1476-4687},
doi={10.1038/s41586-024-07894-z},
url={https://doi.org/10.1038/s41586-024-07894-z}
}

@misc{Hamamci24,
  title={Developing Generalist Foundation Models from a Multimodal Dataset for 3D Computed Tomography}, 
  author={Ibrahim Ethem Hamamci and Sezgin Er and Furkan Almas and Ayse Gulnihan Simsek and Sevval Nil Esirgun and Irem Dogan and Muhammed Furkan Dasdelen and Omer Faruk Durugol and Bastian Wittmann and Tamaz Amiranashvili and Enis Simsar and Mehmet Simsar and Emine Bensu Erdemir and Abdullah Alanbay and Anjany Sekuboyina and Berkan Lafci and Christian Bluethgen and Mehmet Kemal Ozdemir and Bjoern Menze},
  year={2024},
  eprint={2403.17834},
  archivePrefix={arXiv},
  primaryClass={cs.CV},
  url={https://arxiv.org/abs/2403.17834}, 
}

@article{Zhang24survey,
  title={Data-Centric Foundation Models in Computational Healthcare: A Survey},
  author={Zhang, Yunkun and Gao, Jin and Tan, Zheling and Zhou, Lingfeng and Ding, Kexin and Zhou, Mu and Zhang, Shaoting and Wang, Dequan},
  journal={arXiv preprint arXiv:2401.02458},
  year={2024}
}

@article{Chen24survey,
  title={A survey of medical vision-and-language applications and their techniques},
  author={Chen, Qi and Zhao, Ruoshan and Wang, Sinuo and Phan, Vu Minh Hieu and Hengel, Anton van den and Verjans, Johan and Liao, Zhibin and To, Minh-Son and Xia, Yong and Chen, Jian and others},
  journal={arXiv preprint arXiv:2411.12195},
  year={2024}
}

@article{AlDhabyani20,
  title = {Dataset of breast ultrasound images},
  journal = {Data in Brief},
  volume = {28},
  pages = {104863},
  year = {2020},
  issn = {2352-3409},
  doi = {https://doi.org/10.1016/j.dib.2019.104863},
  url = {https://www.sciencedirect.com/science/article/pii/S2352340919312181},
  author = {Walid Al-Dhabyani and Mohammed Gomaa and Hussien Khaled and Aly Fahmy}
}

@ARTICLE{Cai24,
  author={Cai, De and Chen, Jie and Zhao, Junhan and Xue, Yuan and Yang, Sen and Yuan, Wei and Feng, Min and Weng, Haiyan and Liu, Shuguang and Peng, Yulong and Zhu, Junyou and Wang, Kanran and Jackson, Christopher and Tang, Hongping and Huang, Junzhou and Wang, Xiyue},
  journal={IEEE Transactions on Medical Imaging}, 
  title={HiCervix: An Extensive Hierarchical Dataset and Benchmark for Cervical Cytology Classification}, 
  year={2024},
  volume={43},
  number={12},
  pages={4344-4355},
  doi={10.1109/TMI.2024.3419697}
}

@ARTICLE{Fabio16,
  author={Spanhol, Fabio A. and Oliveira, Luiz S. and Petitjean, Caroline and Heutte, Laurent},
  journal={IEEE Transactions on Biomedical Engineering}, 
  title={A Dataset for Breast Cancer Histopathological Image Classification}, 
  year={2016},
  volume={63},
  number={7},
  pages={1455-1462},
  doi={10.1109/TBME.2015.2496264}
}

@inproceedings{Buitinck13,
  author    = {Lars Buitinck and Gilles Louppe and Mathieu Blondel and
                Fabian Pedregosa and Andreas Mueller and Olivier Grisel and
                Vlad Niculae and Peter Prettenhofer and Alexandre Gramfort
                and Jaques Grobler and Robert Layton and Jake VanderPlas and
                Arnaud Joly and Brian Holt and Ga{\"{e}}l Varoquaux},
  title     = {{API} design for machine learning software: experiences from the scikit-learn
                project},
  booktitle = {ECML PKDD Workshop: Languages for Data Mining and Machine Learning},
  year      = {2013},
  pages = {108--122},
}

@inproceedings{SimonyanZ14,
  author    = {Karen Simonyan and Andrew Zisserman},
  editor    = {Yoshua Bengio and Yann LeCun},
  title     = {Very Deep Convolutional Networks for Large-Scale Image Recognition},
  booktitle = {3rd International Conference on Learning Representations, {ICLR} 2015,
               San Diego, CA, USA, May 7-9, 2015, Conference Track Proceedings},
  year      = {2015},
  url       = {http://arxiv.org/abs/1409.1556},
}

@article{Fang24,
  title={Eva-02: A visual representation for neon genesis},
  author={Fang, Yuxin and Sun, Quan and Wang, Xinggang and Huang, Tiejun and Wang, Xinlong and Cao, Yue},
  journal={Image and Vision Computing},
  pages={105171},
  year={2024},
  publisher={Elsevier}
}

@inproceedings{you23,
  title={Cxr-clip: Toward large scale chest x-ray language-image pre-training},
  author={You, Kihyun and Gu, Jawook and Ham, Jiyeon and Park, Beomhee and Kim, Jiho and Hong, Eun K and Baek, Woonhyuk and Roh, Byungseok},
  booktitle={International Conference on Medical Image Computing and Computer-Assisted Intervention},
  pages={101--111},
  year={2023},
  organization={Springer}
}

@misc{Bhuvaji20,
	title={Brain Tumor Classification (MRI)},
	url={https://www.kaggle.com/dsv/1183165},
	DOI={10.34740/KAGGLE/DSV/1183165},
	publisher={Kaggle},
	author={Sartaj Bhuvaji and Ankita Kadam and Prajakta Bhumkar and Sameer Dedge and Swati Kanchan},
	year={2020}
}

@article{Liu22,
title = {DeepDRiD: Diabetic Retinopathy—Grading and Image Quality Estimation Challenge},
journal = {Patterns},
pages = {100512},
year = {2022},
issn = {2666-3899},
doi = {https://doi.org/10.1016/j.patter.2022.100512},
url = {https://www.sciencedirect.com/science/article/pii/S2666389922001040},
author = {Ruhan Liu and Xiangning Wang and Qiang Wu and Ling Dai and Xi Fang and Tao Yan and Jaemin Son and Shiqi Tang and Jiang Li and Zijian Gao and Adrian Galdran and J.M. Poorneshwaran and Hao Liu and Jie Wang and Yerui Chen and Prasanna Porwal and Gavin Siew {Wei Tan} and Xiaokang Yang and Chao Dai and Haitao Song and Mingang Chen and Huating Li and Weiping Jia and Dinggang Shen and Bin Sheng and Ping Zhang},
}

@article{Tschandl18,
  title={The HAM10000 dataset, a large collection of multi-source dermatoscopic images of common pigmented skin lesions},
  author={Tschandl, Philipp and Rosendahl, Cliff and Kittler, Harald},
  journal={Scientific data},
  volume={5},
  number={1},
  pages={1--9},
  year={2018},
  publisher={Nature Publishing Group}
}

@inproceedings{Codella18,
  author={Codella, Noel C. F. and Gutman, David and Celebi, M. Emre and Helba, Brian and Marchetti, Michael A. and Dusza, Stephen W. and Kalloo, Aadi and Liopyris, Konstantinos and Mishra, Nabin and Kittler, Harald and Halpern, Allan},
  booktitle={2018 IEEE 15th International Symposium on Biomedical Imaging (ISBI 2018)}, 
  title={Skin lesion analysis toward melanoma detection: A challenge at the 2017 International symposium on biomedical imaging (ISBI), hosted by the international skin imaging collaboration (ISIC)}, 
  year={2018},
  volume={},
  number={},
  pages={168-172},
  doi={10.1109/ISBI.2018.8363547}}

@article{Hernandez24,
  title={Bcn20000: Dermoscopic lesions in the wild},
  author={Hern{\'a}ndez-P{\'e}rez, Carlos and Combalia, Marc and Podlipnik, Sebastian and Codella, Noel CF and Rotemberg, Veronica and Halpern, Allan C and Reiter, Ofer and Carrera, Cristina and Barreiro, Alicia and Helba, Brian and others},
  journal={Scientific Data},
  volume={11},
  number={1},
  pages={641},
  year={2024},
  publisher={Nature Publishing Group UK London}
}

@article{Rodriguez25,
    title = {A Foundation Language-Image Model of the Retina (FLAIR): encoding expert knowledge in text supervision},
    author = {Julio Silva-Rodr\'{i}guez and Hadi Chakor and Riadh Kobbi and Jose Dolz and Ismail {Ben Ayed}},
    journal = {Medical Image Analysis},
    volume = {99},
    pages = {103357},
    year = {2025},
    issn = {1361-8415},
}

@misc{Oquab23,
  title={DINOv2: Learning Robust Visual Features without Supervision},
  author={Oquab, Maxime and Darcet, Timothée and Moutakanni, Theo and Vo, Huy V. and Szafraniec, Marc and Khalidov, Vasil and Fernandez, Pierre and Haziza, Daniel and Massa, Francisco and El-Nouby, Alaaeldin and Howes, Russell and Huang, Po-Yao and Xu, Hu and Sharma, Vasu and Li, Shang-Wen and Galuba, Wojciech and Rabbat, Mike and Assran, Mido and Ballas, Nicolas and Synnaeve, Gabriel and Misra, Ishan and Jegou, Herve and Mairal, Julien and Labatut, Patrick and Joulin, Armand and Bojanowski, Piotr},
  journal={arXiv:2304.07193},
  year={2023}
}

@misc{Yu22,
  title={CoCa: Contrastive Captioners are Image-Text Foundation Models}, 
  author={Jiahui Yu and Zirui Wang and Vijay Vasudevan and Legg Yeung and Mojtaba Seyedhosseini and Yonghui Wu},
  year={2022},
  eprint={2205.01917},
  archivePrefix={arXiv},
  primaryClass={cs.CV},
  url={https://arxiv.org/abs/2205.01917}, 
}

@article{Weinstein13,
  title={The cancer genome atlas pan-cancer analysis project},
  author={Weinstein, John N and Collisson, Eric A and Mills, Gordon B and Shaw, Kenna R and Ozenberger, Brad A and Ellrott, Kyle and Shmulevich, Ilya and Sander, Chris and Stuart, Joshua M},
  journal={Nature genetics},
  volume={45},
  number={10},
  pages={1113--1120},
  year={2013},
  publisher={Nature Publishing Group}
}

@article{sainburg21,
  title={Parametric UMAP Embeddings for Representation and Semisupervised Learning},
  author={Sainburg, Tim and McInnes, Leland and Gentner, Timothy Q},
  journal={Neural Computation},
  volume={33},
  number={11},
  pages={2881--2907},
  year={2021},
  publisher={MIT Press One Rogers Street, Cambridge, MA 02142-1209, USA journals-info~…}
}

@article{soares20,
  title={SARS-CoV-2 CT-scan dataset: A large dataset of real patients CT scans for SARS-CoV-2 identification},
  author={Soares, Eduardo and Angelov, Plamen and Biaso, Sarah and Froes, Michele Higa and Abe, Daniel Kanda},
  journal={MedRxiv},
  pages={2020--04},
  year={2020},
  publisher={Cold Spring Harbor Laboratory Press}
}

@inproceedings{medmnistv1,
    title = {MedMNIST Classification Decathlon: A Lightweight AutoML Benchmark for Medical Image Analysis},
    author = {Yang, Jiancheng and Shi, Rui and Ni, Bingbing},
    booktitle = {IEEE 18th International Symposium on Biomedical Imaging (ISBI)},
    pages = {191--195},
    year = {2021}
}

@article{medmnistv2,
  title={MedMNIST v2-A large-scale lightweight benchmark for 2D and 3D biomedical image classification},
  author={Yang, Jiancheng and Shi, Rui and Wei, Donglai and Liu, Zequan and Zhao, Lin and Ke, Bilian and Pfister, Hanspeter and Ni, Bingbing},
  journal={Scientific Data},
  volume={10},
  number={1},
  pages={41},
  year={2023},
  publisher={Nature Publishing Group UK London}
}

@article{Kermany18,
  title = {Identifying Medical Diagnoses and Treatable Diseases by Image-Based Deep Learning},
  journal = {Cell},
  volume = {172},
  number = {5},
  pages = {1122-1131.e9},
  year = {2018},
  issn = {0092-8674},
  doi = {https://doi.org/10.1016/j.cell.2018.02.010},
  url = {https://www.sciencedirect.com/science/article/pii/S0092867418301545},
  author = {Daniel S. Kermany and Michael Goldbaum and Wenjia Cai and Carolina C.S. Valentim and Huiying Liang and Sally L. Baxter and Alex McKeown and Ge Yang and Xiaokang Wu and Fangbing Yan and Justin Dong and Made K. Prasadha and Jacqueline Pei and Magdalene Y.L. Ting and Jie Zhu and Christina Li and Sierra Hewett and Jason Dong and Ian Ziyar and Alexander Shi and Runze Zhang and Lianghong Zheng and Rui Hou and William Shi and Xin Fu and Yaou Duan and Viet A.N. Huu and Cindy Wen and Edward D. Zhang and Charlotte L. Zhang and Oulan Li and Xiaobo Wang and Michael A. Singer and Xiaodong Sun and Jie Xu and Ali Tafreshi and M. Anthony Lewis and Huimin Xia and Kang Zhang}}

@article{goh2021multimodal,
  title={Multimodal neurons in artificial neural networks},
  author={Goh, Gabriel and Cammarata, Nick and Voss, Chelsea and Carter, Shan and Petrov, Michael and Schubert, Ludwig and Radford, Alec and Olah, Chris},
  journal={Distill},
  volume={6},
  number={3},
  pages={e30},
  year={2021},
  doi = {10.23915/distill.00030}
}

@article{li2024llava,
  title={LLaVA-NeXT-Interleave: Tackling Multi-image, Video, and 3D in Large Multimodal Models},
  author={Li, Feng and Zhang, Renrui and Zhang, Hao and Zhang, Yuanhan and Li, Bo and Li, Wei and Ma, Zejun and Li, Chunyuan},
  journal={arXiv preprint arXiv:2407.07895},
  year={2024}
}

@inproceedings{wolf-etal-2020-transformers,
    title = "Transformers: State-of-the-Art Natural Language Processing",
    author = "Thomas Wolf and Lysandre Debut and Victor Sanh and Julien Chaumond and Clement Delangue and Anthony Moi and Pierric Cistac and Tim Rault and Rémi Louf and Morgan Funtowicz and Joe Davison and Sam Shleifer and Patrick von Platen and Clara Ma and Yacine Jernite and Julien Plu and Canwen Xu and Teven Le Scao and Sylvain Gugger and Mariama Drame and Quentin Lhoest and Alexander M. Rush",
    booktitle = "Proceedings of the 2020 Conference on Empirical Methods in Natural Language Processing: System Demonstrations",
    month = oct,
    year = "2020",
    address = "Online",
    publisher = "Association for Computational Linguistics",
    url = "https://www.aclweb.org/anthology/2020.emnlp-demos.6",
    pages = "38--45"
}

@article{clark2013cancer,
  title={The Cancer Imaging Archive (TCIA): maintaining and operating a public information repository},
  author={Clark, Kenneth and Vendt, Bruce and Smith, Kirk and Freymann, John and Kirby, Justin and Koppel, Paul and Moore, Stephen and Phillips, Stanley and Maffitt, David and Pringle, Michael and others},
  journal={Journal of digital imaging},
  volume={26},
  number={6},
  pages={1045--1057},
  year={2013},
  publisher={Springer}
}

@misc{Zenodo,
  doi = {10.25495/7GXK-RD71},
  url = {https://www.zenodo.org/},
  author = {{European Organization For Nuclear Research} and {OpenAIRE}},
  language = {en},
  title = {Zenodo},
  publisher = {CERN},
  year = {2013}
}

@misc{kaggle,
title={Kaggle},
howpublished={\url{https://www.kaggle.com}},
}

@article{sellergren2025medgemma,
  title={Medgemma technical report},
  author={Sellergren, Andrew and Kazemzadeh, Sahar and Jaroensri, Tiam and Kiraly, Atilla and Traverse, Madeleine and Kohlberger, Timo and Xu, Shawn and Jamil, Fayaz and Hughes, C{\'\i}an and Lau, Charles and others},
  journal={arXiv preprint arXiv:2507.05201},
  year={2025}
}

@InProceedings{Zhai_2023_ICCV,
    author    = {Zhai, Xiaohua and Mustafa, Basil and Kolesnikov, Alexander and Beyer, Lucas},
    title     = {Sigmoid Loss for Language Image Pre-Training},
    booktitle = {Proceedings of the IEEE/CVF International Conference on Computer Vision (ICCV)},
    month     = {October},
    year      = {2023},
    pages     = {11975-11986}
}

@article{wang2022transformer,
  title={Transformer-based unsupervised contrastive learning for histopathological image classification},
  author={Wang, Xiyue and Yang, Sen and Zhang, Jun and Wang, Minghui and Zhang, Jing and Yang, Wei and Huang, Junzhou and Han, Xiao},
  journal={Medical image analysis},
  volume={81},
  pages={102559},
  year={2022},
  publisher={Elsevier}
}

@InProceedings{Liu_2021_ICCV,
    author    = {Liu, Ze and Lin, Yutong and Cao, Yue and Hu, Han and Wei, Yixuan and Zhang, Zheng and Lin, Stephen and Guo, Baining},
    title     = {Swin Transformer: Hierarchical Vision Transformer Using Shifted Windows},
    booktitle = {Proceedings of the IEEE/CVF International Conference on Computer Vision (ICCV)},
    month     = {October},
    year      = {2021},
    pages     = {10012-10022}
}

@inproceedings{yu2024clip,
  title={CLIP-DR: Textual knowledge-guided diabetic retinopathy grading with ranking-aware prompting},
  author={Yu, Qinkai and Xie, Jianyang and Nguyen, Anh and Zhao, He and Zhang, Jiong and Fu, Huazhu and Zhao, Yitian and Zheng, Yalin and Meng, Yanda},
  booktitle={International Conference on Medical Image Computing and Computer-Assisted Intervention},
  pages={667--677},
  year={2024},
  organization={Springer}
}

@article{silva2025foundation,
  title={A foundation language-image model of the retina (flair): Encoding expert knowledge in text supervision},
  author={Silva-Rodriguez, Julio and Chakor, Hadi and Kobbi, Riadh and Dolz, Jose and Ayed, Ismail Ben},
  journal={Medical Image Analysis},
  volume={99},
  pages={103357},
  year={2025},
  publisher={Elsevier}
}

@inproceedings{lin2023pmc,
  title={Pmc-clip: Contrastive language-image pre-training using biomedical documents},
  author={Lin, Weixiong and Zhao, Ziheng and Zhang, Xiaoman and Wu, Chaoyi and Zhang, Ya and Wang, Yanfeng and Xie, Weidi},
  booktitle={International Conference on Medical Image Computing and Computer-Assisted Intervention},
  pages={525--536},
  year={2023},
  organization={Springer}
}

@article{ruckert2024rocov2,
  title={Rocov2: Radiology objects in context version 2, an updated multimodal image dataset},
  author={R{\"u}ckert, Johannes and Bloch, Louise and Br{\"u}ngel, Raphael and Idrissi-Yaghir, Ahmad and Sch{\"a}fer, Henning and Schmidt, Cynthia S and Koitka, Sven and Pelka, Obioma and Abacha, Asma Ben and G. Seco de Herrera, Alba and others},
  journal={Scientific Data},
  volume={11},
  number={1},
  pages={688},
  year={2024},
  publisher={Nature Publishing Group UK London}
}

@article{subramanian2020medicat,
  title={Medicat: A dataset of medical images, captions, and textual references},
  author={Subramanian, Sanjay and Wang, Lucy Lu and Mehta, Sachin and Bogin, Ben and Van Zuylen, Madeleine and Parasa, Sravanthi and Singh, Sameer and Gardner, Matt and Hajishirzi, Hannaneh},
  year={2020},
  booktitle={Findings of EMNLP},
}

@inproceedings{baghbanzadeh2025advancing,
  title={Advancing medical representation learning through high-quality data},
  author={Baghbanzadeh, Negin and Fallahpour, Adibvafa and Parhizkar, Yasaman and Ogidi, Franklin and Roy, Shuvendu and Ashkezari, Sajad and Khazaie, Vahid Reza and Colacci, Michael and Etemad, Ali and Afkanpour, Arash and others},
  booktitle={International Conference on Medical Image Computing and Computer-Assisted Intervention},
  pages={24--33},
  year={2025},
  organization={Springer}
}

@article{li2024gmai,
  title={Gmai-vl \& gmai-vl-5.5 m: A large vision-language model and a comprehensive multimodal dataset towards general medical ai},
  author={Li, Tianbin and Su, Yanzhou and Li, Wei and Fu, Bin and Chen, Zhe and Huang, Ziyan and Wang, Guoan and Ma, Chenglong and Chen, Ying and Hu, Ming and others},
  journal={arXiv preprint arXiv:2411.14522},
  year={2024}
}

@article{Li24,
  title={VisionUnite: A Vision-Language Foundation Model for Ophthalmology Enhanced with Clinical Knowledge},
  author={Li, Zihan and Song, Diping and Yang, Zefeng and Wang, Deming and Li, Fei and Zhang, Xiulan and Kinahan, Paul E and Qiao, Yu},
  journal={IEEE Transactions on Pattern Analysis and Machine Intelligence},
  year={2025},
  publisher={IEEE}
}

@article{lu2025integrating,
  title={Integrating language into medical visual recognition and reasoning: A survey},
  author={Lu, Yinbin and Wang, Alan},
  journal={Medical Image Analysis},
  pages={103514},
  year={2025},
  publisher={Elsevier}
}

@article{liu2017current,
  title={The current role of image compression standards in medical imaging},
  author={Liu, Feng and Hernandez-Cabronero, Miguel and Sanchez, Victor and Marcellin, Michael W and Bilgin, Ali},
  journal={Information},
  volume={8},
  number={4},
  pages={131},
  year={2017},
  publisher={MDPI}
}

@article{koff2013evaluation,
  title={Evaluation of irreversible compression ratios for medical images thin slice CT and update of Canadian Association of Radiologists (CAR) guidelines},
  author={Koff, David and Bak, Peter and Matos, Andr{\'e} and Norman, Geoff},
  journal={Journal of digital imaging},
  volume={26},
  number={3},
  pages={440--446},
  year={2013},
  publisher={Springer}
}

@article{herath2025systematic,
  title={A Systematic Review of Medical Image Quality Assessment},
  author={Herath, HMSS and Herath, HMKKMB and Madusanka, Nuwan and Lee, Byeong-Il},
  journal={Journal of Imaging},
  volume={11},
  number={4},
  pages={100},
  year={2025}
}

\end{document}